\documentclass[Preprint,12pt]{elsarticle}
\usepackage{times}
\usepackage[utf8]{inputenc}
\usepackage[usenames, dvipsnames]{color}
\usepackage{booktabs}
\usepackage[linesnumbered,ruled]{algorithm2e}
\usepackage{graphicx,amssymb,mathrsfs,amsmath,pifont,amscd,latexsym,amsfonts,amsthm}
\usepackage{epstopdf}
\usepackage{float}
\usepackage[scriptsize]{subfigure}
\usepackage{epsfig}
\usepackage{multirow}
\usepackage[switch]{lineno}
\usepackage{caption}
\usepackage[colorlinks,linkcolor=red,anchorcolor=blue,citecolor=green]{hyperref}
\journal{ISPRS Journal of Photogrammetry and Remote Sensing}

\newcommand{\norm}[1]{\lVert#1\rVert}

\usepackage{numcompress}

\begin{document}

\begin{frontmatter}
\title{X-ModalNet: A Semi-Supervised Deep Cross-Modal Network for Classification of Remote Sensing Data}


\author{
Danfeng Hong\textsuperscript{a,b}, Naoto Yokoya\textsuperscript{c,d}, Gui-Song Xia\textsuperscript{e,f,g}, Jocelyn Chanussot\textsuperscript{h,i}, Xiao Xiang Zhu\textsuperscript{a,b}}

\address{
	\textsuperscript{a}Remote Sensing Technology Institute, German Aerospace Center, 82234 Wessling, Germany.\\
	\textsuperscript{b}Signal Processing in Earth Observation, Technical University of Munich, 80333 Munich, Germany.\\
	\textsuperscript{c}Graduate School of Frontier Sciences, The University of Tokyo, 277-8561 Chiba, Japan.\\
	\textsuperscript{d}Geoinformatics Unit, RIKEN Center for Advanced Intelligence Project, RIKEN, 103-0027 Tokyo, Japan.\\
	\textsuperscript{e}School of Computer Science, Wuhan University, 430072 Wuhan, China\\
	\textsuperscript{f} Institute of Artificial Intelligence, Wuhan University, 430072 Wuhan, China\\
	\textsuperscript{g}State Key Laboratory for Information Engineering in Surveying, Mapping and Remote Sensing, Wuhan University, 430079 Wuhan, China\\
	\textsuperscript{h}Univ. Grenoble Alpes, INRIA, CNRS, Grenoble INP, LJK, 38000 Grenoble, France\\
	\textsuperscript{i}Aerospace Information Research Institute, Chinese Academy of Sciences, 100094 Beijing, China.\\
}
\begin{abstract}
This paper addresses the problem of semi-supervised transfer learning with limited cross-modality data in remote sensing. A large amount of multi-modal earth observation images, such as multispectral imagery (MSI) or synthetic aperture radar (SAR) data, are openly available on a global scale, enabling parsing global urban scenes through remote sensing imagery. However, their ability in identifying materials (pixel-wise classification) remains limited, due to the noisy collection environment and poor discriminative information as well as limited number of well-annotated training images. To this end, we propose a novel cross-modal deep-learning framework, called X-ModalNet, with three well-designed modules: self-adversarial module, interactive learning module, and label propagation module, by learning to transfer more discriminative information from a small-scale hyperspectral image (HSI) into the classification task using a large-scale MSI or SAR data. Significantly, X-ModalNet generalizes well, owing to propagating labels on an updatable graph constructed by high-level features on the top of the network, yielding semi-supervised cross-modality learning. We evaluate X-ModalNet on two multi-modal remote sensing datasets (HSI-MSI and HSI-SAR) and achieve a significant improvement in comparison with several state-of-the-art methods.
\end{abstract}

\end{frontmatter}

\graphicspath{{figures/}}
\section{Introduction}

Currently operational radar (e.g., Sentinel-1) and optical broadband (multispectral) satellites (e.g., Sentinel-2 and Landsat-8) enable the synthetic aperture radar (SAR) \cite{kang2020learning} and multispectral image (MSI) \cite{zhang2019estimation} openly available on a global scale. Therefore, there has been a growing interest in understanding our environment through remote sensing (RS) images, which is of great benefit to many potential applications such as image classification \citep{tuia2015multiclass,han2018edge,srivastava2019understanding,cao2020an}, object and change detection \citep{zhang2018change,wu2019orsim,zhang2019detecting,wu2019fourier}, mineral exploration \citep{gao2017new,hong2018sulora,hong2019augmented,yao2019nonconvex}, multi-modality data analysis \citep{hong2019learnable,hu2019comparative,yang2019introduction,hong2020learning}, to name a few. In particular, RS data classification is a fundamental but still challenging problem across computer vision and RS fields. It aims to assign a semantic category to each pixel in a studied urban scene. For example, in \cite{gao2017optimized}, spectral-spatial information is applied to significantly suppress the influence of noise in dimensionality reduction, and the proposed method is obviously effective in extracting nonlinear features and improving the classification accuracy.

\begin{figure*}[!t]
\centering
\includegraphics[width=0.8\textwidth]{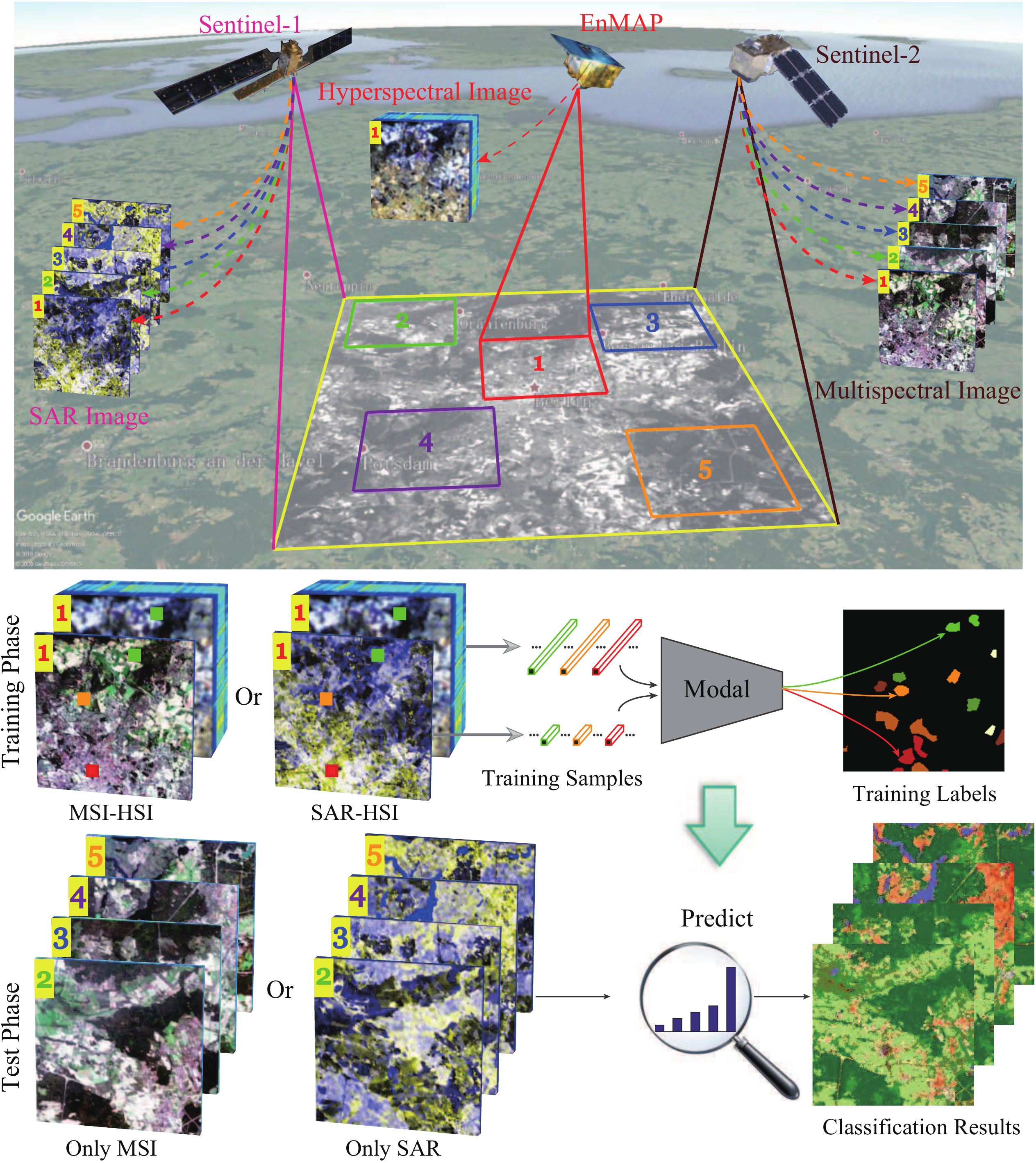}
\caption{Our proposed solution (bottom) for the cross-modality learning problem in RS (top). \textit{Top}: Given a large-scale urban area in \textcolor{yellow}{yellow}, both SAR in \textcolor{magenta}{magenta} and MSI in \textcolor[rgb]{0.5,0,0}{chestnut} are openly and largely available with a high spatial resolution but limited by poor feature discrimination power, while the HSI in \textcolor{red}{red} is information-rich but only a small-scale available, as shown in area 1 overlapped with SAR (or MSI). \textit{Bottom}: The model is trained on multimodalities (e.g., HSI-MSI or HSI-SAR) with the sparse training labels, and one modality is absent in the process of predicting.}
\label{fig:motivation}
\end{figure*}

Recently, enormous efforts have been made on developing deep learning (DL)-based approaches \citep{lecun2015deep}, such as deep neural networks (DNNs) and convolutional neural networks (CNNs), to parse urban scenes by using street view images. Yet it is less investigated at the level of satellite-borne or aerial images. Bridging advanced learning-based techniques or vision algorithms with RS imagery could allow for a variety of new applications potentially conducted on a larger and even a global scale. A qualitative comparison is given in Table \ref{table:ClarifySimAndDiff} to highlight the differences as well as advantages and disadvantages in the classification task using different scene images (e.g., street view or RS images).

\subsection{Motivation and Objective}
We clarify our motivation to answer the following three ``why'' questions: 1) \textit{Why classify or parse RS images?} 2) \textit{Why use multimodal data?} 3) \textit{Why learn the cross-modal representation?}

\begin{table*}[!t]
\centering
\caption{Qualitative comparison of urban scene parsing using street view images and RS images in terms of goal, acquisition perspective, scene covering scale, spatial resolution, feature diversity, data accessibility, and ground truth maps used for training.}
\resizebox{0.5\textwidth}{!}{
\begin{tabular}{lcc}
\toprule[1.5pt]
Urban Scene Parsing & Street View Images & RS Images\\
\hline\hline
Goal & \multicolumn{2}{c}{Pixel-wise Classification}\\
Perspective & Horizontal & ``Bird's''\\
Scene Scale & Small & Large \\
Spatial Resolution & High & Low \\
Feature Diversity & Low & High \\
Accessibility & Moderate & Easy \\
Ground Truth Maps & Dense & Sparse \\
\bottomrule[1.5pt]
\end{tabular}
}
\label{table:ClarifySimAndDiff}
\end{table*}

\begin{itemize}
\item{\bf From Street to Earth Vision} \;
Remotely sensed imagery can provide a new insight for global urban scene understanding. The data in Earth Vision, on one hand, benefit from a ``bird's perspective,'' providing a structure-related multiview surface information; and, on the other hand, it is acquired on a wider and even global scale.
\item{\bf From Unimodal to Multimodal Data} \;
Limited by the low image resolution and a handful of labeled samples, unimodal RS data are inevitable to meet the bottleneck in performance gain, despite being able to be openly and largely acquired. Therefore, an alternative to maximize the classification accuracy is to jointly leverage the multimodal data.
\item{\bf From Multimodal to Crossmodal Learning} \;
In reality, a large amount of information-rich data, such as hyperspectral imagery (HSI), are hardly collected due to technical limitations of satellite sensors. Thus, only the limited multimodal correspondences can be used to train a model, while one modality is absent in the test phase. This is a typical cross-modality learning (CML) issue.
\end{itemize}

Fig. \ref{fig:motivation} illustrates the to-be-solved problem and potential solution, where MSI in magenta or SAR in cyan is freely available at a large and even global scale but they are limited by relatively poor feature representation ability, while the HSI in red is characterized by rich spectral information but fails to be acquired in a large-covered area. This naturally leads to a general but interesting question: \textit{can a limited amount of spectrally discriminative HSI improve the parsing performance of a large amount of low-quality data (SAR or MSI) in the large-scale classification or mapping task?} A feasible solution to the problem is the CML.

Motivated by the above analysis, the CML issue that we aim at tackling can be further generalized to three specific challenges related to computer vision or machine learning.

\begin{itemize}
\item{RS images acquired from the satellites or airplanes inevitably suffer from various variations caused by environmental conditions (e.g., illumination and topology changes, atmospheric effects)  and instrumental configurations (e.g., sensor noise).}
\item{Multimodal RS data are usually characterized by the different properties. Blending multi / cross-modal representation in a more effective and compacted way is still an important challenge in our case.}
\item{RS images in Earth Vision can provide a larger-scale visual field. This tends to lead to costly labeling and noisy annotations in the process of data preparation.}
\end{itemize}
According to the three factors, our objective can be summarized to develop novel approaches or improve the existing ones, yielding a more discriminative multimodality blending and robust against various variabilities in RS images with the limited number of training annotations.

\subsection{Method Overview and Contributions}
Towards the aforementioned goals, a novel cross-modal DL framework is proposed in a semi-supervised fashion, called X-ModalNet, for RS image classification. As illustrated in Fig. \ref{fig:Workflow}, a three-stream network is developed to learn the multimodal joint representation in consideration of unlabeled samples, where the network parameters would be shared from the same modalities. Moreover, an interactive learning strategy is modeled across the two modalities to facilitate the information blending more effectively. Prior to the interactive learning ({IL}) module, we also embed a self-adversarial ({SA}) module robustly against noise attack, thereby enhance the model's generalization capability. To fully make use of unlabeled samples, we iteratively update pseudo-labels by label propagation ({LP}) on the graph constructed by high-level hidden representations. Extensive experiments are conducted on two multimodal datasets (HSI-MSI and HSI-SAR), showing the effectiveness and superiority of our proposed X-ModalNet in the RS data classification task.

The main contributions can be highlighted in four-folds:
\begin{itemize}
\item To our best knowledge, this is the first time to investigate the HSI-aided CML's case by designing such deep cross-modal network (X-ModalNet) in RS fields for improving the classification accuracy of only using MSI or SAR with the aid of a limited amount of HSI samples.
\item According to spatially high resolution of MSI (SAR) as well as spectrally high resolution of HSI, our X-ModalNet is a novel and promising network architecture reasonably, which takes a hybrid network as backbone, that is, CNN for MSI or SAR and DNN for HSI. Such design enables the best full use of high spatial and rich spectral information from MSI or SAR and HSI, respectively.
\item We propose two novel plug-and-play modules: {SA} module and {IL} module, aiming at improving the robustness and discrimination of the multimodal representation. On the one hand, we modularize the idea of generative adversarial networks (GANs) \citep{goodfellow2014generative} into the network to generate robust feature representations by simultaneously learning original features and adversarial features in SA module. On the other hand, we design the IL module for better information blending across modalities by interactively sharing the network weights to generate more discriminative and compact features.
\item We design an updatable LP mechanism into our proposed end-to-end networks by progressively optimizing pseudo-labels to further find a better decision boundary.
\item We validate the superiority and effectiveness of X-ModalNet on two cross-modal datasets with extensive ablation analysis, where we collected and processed the Sentinel-1 SAR data for the second datasets.
\end{itemize}

\section{Related Work}
\subsection{Scene Parsing}
Most recently, the research on scene parsing has made unprecedented progress, owing to the powerful DNNs \citep{krizhevsky2012imagenet}. Most of these state-of-the-art DL-based frameworks for scene parsing \citep{yu2015multi, noh2015learning, xia2016zoom, zhao2017pyramid,li2017foveanet, zhang2018context, chen2018deeplab, rasti2020feature} are closely associated with two seminal works presented on the prototype of deep CNN: fully convolutional network \citep{long2015fully}, DeepLab \citep{chen2018deeplab}.  However, a nearly horizontal field of vision makes it difficult to parse a large urban area without extremely diverse training samples. Therefore, RS images might be a feasible and desirable alternative.

We observed that the RS imagery has attracted increasing interest in computer vision field \citep{lanaras2015hyperspectral, xia2018dota, marcos2018learning}, as it generally holds a diversified and structured source of information, which can be used for better scene understanding and further make a significant breakthrough in global urban motoring and planning \citep{demir2018deepglobe}. Chen \textit{et al.} \citep{chen2014deep} fed the vector-based input into a DNN for predicting the category labels in the HSI. They extended their work by training a CNN to achieve a spatial-spectral HSI classification CNN \citep{chen2016deep}. Hang \textit{et al.} \citep{hang2019cascaded} utilized a Cascaded RNN to parse the HSI scenes. Perceptibly, the scene parsing in Earth Vision is normally performed by training an end-to-end network with a vector-based or a patch-based input, as the sparse labels (see Fig. \ref{fig:motivation}) can not support us to train a FCN-like model. As listed in Table \ref{table:ClarifySimAndDiff}, RS images are noisy but low resolution, and are relatively expensive and time-consuming in labeling, limiting the performance improvement. A feasible solution to the issue is to introduce other modalities (e.g., HSI) with more discriminative information, yielding multimodal data analysis.

\subsection{Multi/Cross-Modal Learning}
Multimodal representation learning related to DNN can be categorized into two aspects \citep{baltruvsaitis2018multimodal}.

\subsubsection{Joint Representation Learning}
The basic idea is to find a joint space where the discriminative feature representation is expected to be learned over multi-modalities with multilayered neural networks. Although some recent works have attempted to challenge the CML issue by using joint representation learning strategy, e.g., \cite{hong2019cospace,hong2020learning}, yet these methods remain limited in data representation and fusion, particularly for heterogeneous data, due to their linearized modeling. A representative work in the multimodal deep learning (MDL) was proposed by Ngiam \textit{et al.} \citep{ngiam2011multimodal}, in which the high-level features for each modality are extracted using a stacked denoising autoencoder (SDAE) and then jointly learned to a multimodal representation by an additional encoder layer. \citep{silberer2014learning} extended the work to a semi-supervised version by additionally using a term into loss function that predicts the labels. Similarly, Srivastava \textit{et al.} utilized the deep belief network \citep{srivastava2012learning} and deep Boltzmann machines \citep{srivastava2012multimodal} to explain the multimodal data fusion or learning from the perspective of probabilistic graphical models. In \citep{rastegar2016mdl}, a novel multimodal DL with cross weights (MDL-CW) is proposed to interactively represent the multimodal features for a more effective information blending. Besides, some follow-up work has been successively proposed to learn the joint feature representation more effectively and efficiently \citep{ouyang2014multi, wang2014effective, peng2016cross, silberer2017visually, luo2017label, liu2019stfnet}.

\subsubsection{Coordinated Representation Learning}
It builds the disjunct subnetworks to learn the discriminative features independently for each modality and couples them by enforcing various structured constraints onto the resulting encoder layers. These structures can be measured by similarity \citep{frome2013devise, feng2014cross},  correlation \citep{chandar2016correlational}, and sequentiality \citep{vendrov2015order}, etc.

In recent years, some tentative work has been proposed for multimodal data analysis in RS \citep{gomez2015multimodal, kampffmeyer2016semantic, mattyus2016hd, audebert2016semantic, audebert2017joint, zampieri2018multimodal, ghosh2018stacked}. Related to ours for scene parsing with multimodal deep networks, an early deep fusion architecture, simply stacking all multi-modalities as input, is used for semantic segmentation of urban RS images \citep{kampffmeyer2016semantic}. In \citep{audebert2017joint}, optical and OpenStreetMap \citep{haklay2008openstreetmap} data are jointly applied with a two-stream deep network for getting a faster and better semantic map. Audebert \textit{et al.} \citep{audebert2018beyond} parsed the urban scenes under the SegNet-like architecture \citep{badrinarayanan2017segnet} by using MSI and Lidar. Similarly, Ghosh \textit{et al.} \citep{ghosh2018stacked} proposed a stacked U-Nets for material segmentation of RS imagery. Nevertheless, these methods are mostly developed with optical (MSI or RGB) or Lidar data for the rough-grained scene parsing (only few categories) and fail to perform sufficiently well in a complex urban scene due to the relatively poor feature representation ability behind the networks, especially in CML \cite{ngiam2011multimodal}.

\subsection{Semi-Supervised Learning}
Considering the fact that the labeling cost is very expensive, particularly for RS images, the use of unlabeled samples has gathered increasing attention as a feasible solution to further improve the classification performance of RS data. There have been many non-DL-based semi-supervised learning approaches in a variety of RS-related applications, such as regression-based multitask learning \cite{hong2019learning,hong2019regression}, manifold alignment \cite{tuia2014semisupervised,Hu2019mima}, factor analysis \cite{zhao2019semi}. Yet this topic is less investigated by using the DL-based approaches. Cao \textit{et al.} \cite{cao2020hyperspectral} integrated CNNs and active learning to better utilize the unlabeled samples for hyperspectral image classification. Riese \textit{et al.} \cite{riese2020supervised} developed a semi-supervised shallow network -- self-organizing map framework -- to classify and estimate physical parameters from MSI and HSI. Nevertheless, how to embed the semi-supervised techniques into deep networks more effectively remains challenging.

\begin{figure*}[!t]
\centering
\includegraphics[width=1\textwidth]{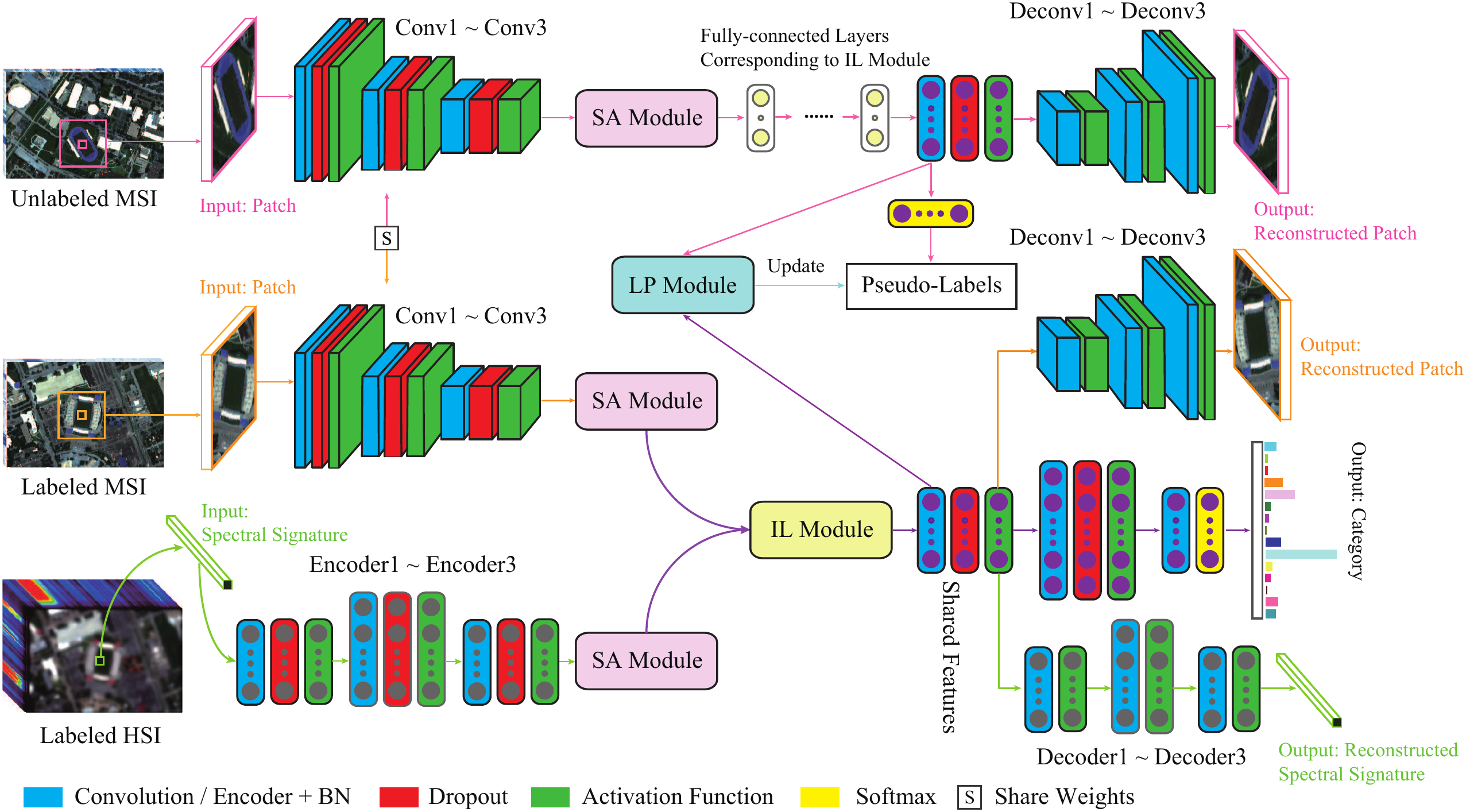}
\caption{An overview of the proposed X-ModalNet. It mainly consists of three modules: (a) {SA} module, (b) {IL} module, and (c) {LP} module, installed in a hybrid (MSI or SAR: CNN and HSI: DNN) semi-supervised multimodal DL framework.}
\label{fig:Workflow}
\end{figure*}

\section{The Proposed X-ModalNet}
The CML's problem setting drives us to develop a robust and discriminative network for pixel-wise classification of RS images in complex scenes. Fig. \ref{fig:Workflow} illustrates the architecture overview of the X-ModalNet, which is built upon a three-stream deep architecture. The \textit{IL} module is designed for highly compact feature blending before feeding the features of each modality into joint representation, and we also equip with the \textit{SA} module and an iterative LP mechanism to improve the robustness and the generalization ability of the proposed X-ModalNet, particularly in the presence of noisy samples.

\subsection{Network Architecture}
The bimodal deep autoencoder (DAE) in \citep{ngiam2011multimodal} is a well-known work in MDL, and we advance it to the proposed X-ModalNet for classification of RS imagery. The differences and improvements mainly lie in four aspects.

\begin{figure*}[!t]
	  \centering
		\subfigure[Self-Adversarial (SA) module]{
			\includegraphics[height=0.2\textheight]{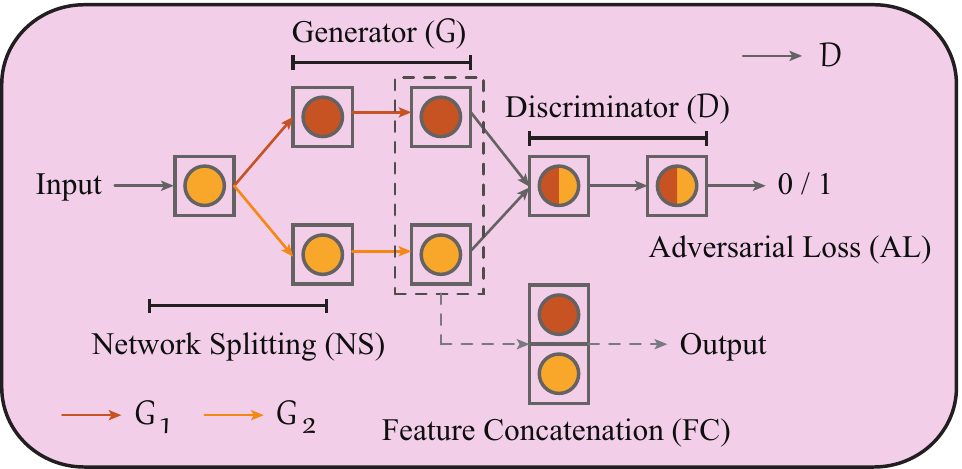}
    		}
        \subfigure[Interactive learning (IL) module]{
			\includegraphics[height=0.2\textheight]{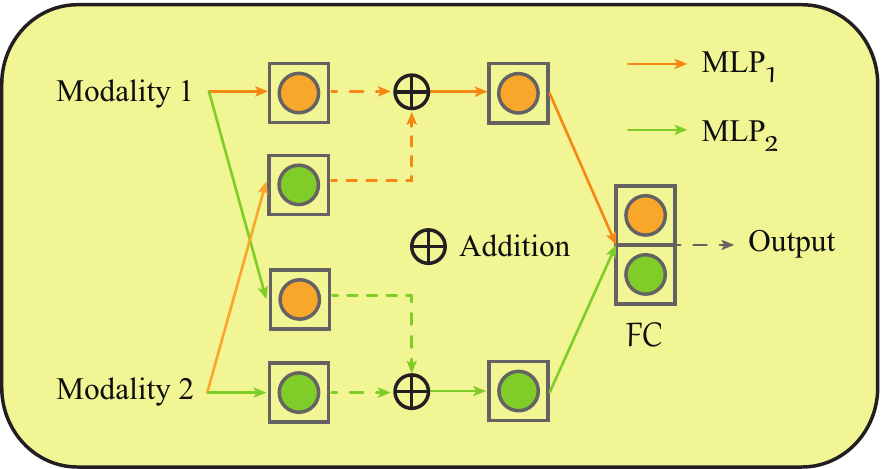}
		}
		\subfigure[Label propagation (LP) module]{
			\includegraphics[height=0.2\textheight]{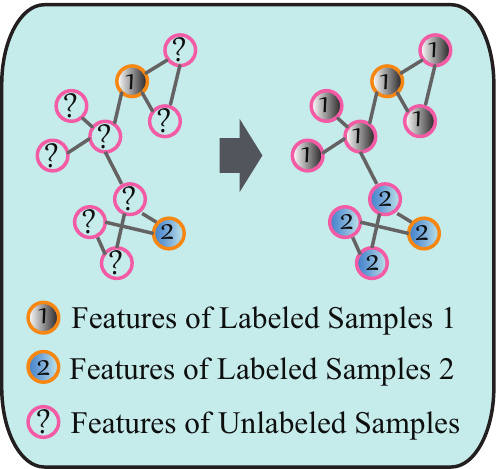}
		}
        \caption{An illustration for three proposed modules in the X-ModalNet: (a) SA module, (b) IL module, and (c) LP module. The arrowed solid lines denote the to-be-learned parameters, and their colors mean the different streams in (a) or modalities in (b). Note that MLP is the abbreviation of multi-layer perception \cite{pal1992multilayer}. For example to see modality 1 in (b), modality 1 reaches the hidden layer (\textcolor{Orange}{orange}) through the parameters and meanwhile modality 2 reaches the hidden layer (\textcolor{green}{green}) through the same parameters.}
\label{fig:modules}
\end{figure*}

\subsubsection{Hybrid Network Architecture}
Similarly to \citep{cangea2017xflow}, we propose a hybrid-stream network architecture in a bimodal DAE fashion, including two CNN-streams on the labeled MSI (SAR) and unlabeled one, and a DNN-stream on HSI, to exploit high spatial information of MSI/SAR data and high spectral information of HSI more effectively. Since hyperspectral imaging enables discrimination between spectrally similar classes (high-spectral resolution) but its swath width from space is narrow compared to multispectral or SAR ones (high-spatial resolution). More specifically, we take the patches centered by pixels as the input of CNN-streams for labeled and unlabeled MSIs (SARs), and the spectral signatures of the corresponding pixels as the input of DNN-stream for labeled HSI. Moreover, the reconstructed patches (CNN-streams) and spectral signatures (DNN-stream) of all pixels as well as the one-hot encoded labels can be regarded as the network outputs.

\subsubsection{Self-Adversarial Module}
Due to the environmental factors (e.g., illumination, physically and chemically atmospheric effects) and instrumental errors, it is inevitable to have some distortions in RS imaging. These noisy images tend to generate attacked samples, thereby hurting the network performance \citep{szegedy2013intriguing, goodfellow2014explaining, melis2017deep}. Unlike the previous adversarial training approaches \citep{donahue2016adversarial, biggio2018wild} that generate adversarial samples in the first place and then feed them into a new network for training, we learn the adversarial information in the feature-based level rather than the sample-based one, with an end-to-end learning process. This might lead to a more robust feature representation in accord with the learned network parameters. As illustrated in Fig. \ref{fig:modules}(a), given a vector-based feature input of the module, the network is first split into two streams ({NS}). It is well-known that the discriminator in GANs enables the generation of adversarial examples to fool the networks. Inspired by it, we assume that in our SA module, one stream extracts or generates the high-level features of the input, while another one correspondingly learns the adversarial features by allowing for an adversarial loss on the top layer ({AL}). In this process, the discriminator can be well regarded as a constraint to achieve the function. In addition, this has been also proven to be effective by the reference \cite{yu2019attributing} to a great extent. Finally, the features represented from the two subnetworks are concatenated as the module output ({FC}) in order to generate more robust feature representations by simultaneously considering the original features and its adversarial features into the network training. Moreover, the superiority of our SA module mainly lies in that the parameters in the module is an end-to-end trainable in the whole X-ModalNet, which can make the learned adversarial features more suitable for our classification tasks. By contrary, if we select to first generate adversarial samples by using an independent GAN and feed them into the classification network together with existing real samples, then the generated adversarial samples could bring the uncertainty for the classification performance improvement. The main reason is that the adversarial samples are generated by an independent GAN, which might be applicable to the GAN but might not be applicable to the classification network because they are trained separately.

\subsubsection{Interactive Learning Module}
We found that in the layer of multimodal joint representation, massive connections occur in variables from the same modality but few neurons across the modalities are activated, even if each modality passes through multiple individual hidden layers before being fed into the joint layer. Different from the \textit{hard-interactive mapping learning} in \citep{rastegar2016mdl,nie2018mutual} that additionally learns the weights across the different modalities, we propose a \textit{soft-interactive learning strategy} that directly copies the weights learned from one modality to another one without additional computational cost and information loss, then fuses them on the top layer only with a simple addition operation, as illustrated in Fig. \ref{fig:modules}(b). This would be capable of learning the inter-modality corrections both effectively and efficiently by reducing the gap between the modalities, yielding a smooth multi-stream networks blending.

\subsubsection{Label Propagation Module}
Beyond the supervised learning, we also consider the unlabeled samples by incorporating the label propagation (see Fig. \ref{fig:modules}(c)) into the networks to further improve the model's generalization. The main workflow in the LP module is detailed as follows:

\begin{itemize}
    \item We first train a classifier on the training set (SVMs used in our case) and predict unlabeled samples by using the trained classifier. These predicted results (pseudo-labels) can be regarded as the network ground truth of unlabeled data stream, which is further considered with real labels into the network training for a multitask learning.
    \item Next, we start to train our networks until convergence occurs. We call this process as one-round network training. Once one-round network training has been completed, the high-level features extracted from the top of the network (see Fig. \ref{fig:Workflow}) are used to update the pseudo-labels using the graph-based LP \cite{zhu2005semi}. The LP algorithm consists of the following two steps.
    \begin{itemize}
        \item Step 1: construct similarity matrix. The similarity matrix $\mathbf{S}$ between any two samples \cite{hong2015novel}, e.g., $x_{i}$ and $x_j$, either labeled or unlabeled, is computed by
        \begin{equation}
        \label{eq1}
        \begin{aligned}
              \mathbf{S}_{i,j} = \mathrm{exp}(-\frac{\norm{x_{i}-x_{j}}^{2}}{\sigma^{2}}),
        \end{aligned}
        \end{equation}
        where $\sigma$ is a hyperparameter determined from the range of $[0.001, 0.01,\\ 0.1, 1, 10, 100]$ by cross-validation on the training set.
        \item Step 2: propagate labels over all samples. Before carrying out LP, a label transfer matrix ($\mathbf{P}$), e.g., from the sample $i$ to the sample $j$, is defined as
        \begin{equation}
        \label{eq2}
        \begin{aligned}
              \mathbf{P}_{i,j} :=\mathbf{P}(i\to j)= \frac{\mathbf{S}_{i,j}}{\sum_{k=1}^{N}\mathbf{S}_{i,k}},
        \end{aligned}
        \end{equation}
        where $N$ is the number of samples. Assume that given $M$ labeled and $N-M$ unlabeled samples with $C$ categories, a soft label matrix $\mathbf{Y}\in \mathbb{R}^{N\times C}$ is constructed, which consists of a labeled matrix $\mathbf{Y}_{l}\in \mathbb{R}^{M\times C}$ and a unlabeled matrix $\mathbf{Y}_{u}\in \mathbb{R}^{(N-M)\times C}$ obtained by one-hot encoding. Our goal is to update the matrix $\mathbf{Y}$, we then have the update rule in the $t$-th ($t\geq 1$) iteration as follows: 1) update $\mathbf{Y}^{t}$ by $\mathbf{P}\mathbf{Y}^{t-1}$; 2) reset $\mathbf{Y}_{l}^{t}$ in $\mathbf{Y}^{t}$ using the original $\mathbf{Y}_{l}$ as $\mathbf{Y}_{l}^{t}=\mathbf{Y}_{l}$; 3) repeat the steps 1) and 2) until convergence.
    \end{itemize}

    We re-feed these updated pseudo-labels, i.e., $\mathbf{Y}_{u}$ into the next-round network training. The workflow is run repeatedly until the pseudo-labels are not changed any more. Note that we experimentally found that three to four repetitions are usually enough, leading to the model convergence.
\end{itemize}

\subsection{Objective Function}
Let $\mathbf{x}_{SA}$ and $\mathbf{z}_{SA}$ be the input and output of the SA module, and then we have
\begin{equation}
\label{eq1}
\begin{aligned}
      &\mathbf{z}_{SA}^{1} = G_1(\mathbf{x}_{SA}),\\
      &\mathbf{z}_{SA}^{2} = G_2(\mathbf{x}_{SA}), \\
      &\mathbf{z}_{SA}=\left[\mathbf{z}_{SA}^{1}, \;  \mathbf{z}_{SA}^{2}\right],
\end{aligned}
\end{equation}
where $G$ is the generative subnetwork that consists of several encoder, normalization (BN) \citep{ioffe2015batch} and dropout  \citep{srivastava2014dropout} layers (see Fig. \ref{fig:modules}). Given the inputs of two modalities $\mathbf{x}_{IL}^{1}$ and $\mathbf{x}_{IL}^{2}$ in the \textit{IL} module, its output ($\mathbf{z}_{IL}$) can be formulated by
\begin{equation}
\label{eq2}
\begin{aligned}
      &\mathbf{z}_{IL}^{11} = MLP_{1}(\mathbf{x}_{IL}^{1}), \\ &\mathbf{z}_{IL}^{22} = MLP_{2}(\mathbf{x}_{IL}^{2}), \\
      &\mathbf{z}_{IL}^{12} = MLP_{2}(\mathbf{x}_{IL}^{1}), \\ &\mathbf{z}_{IL}^{21} = MLP_{1}(\mathbf{x}_{IL}^{2}), \\
      &\mathbf{z}_{IL}=\left[\mathbf{z}_{IL}^{11}+\mathbf{z}_{IL}^{12}, \;  \mathbf{z}_{IL}^{22}+\mathbf{z}_{IL}^{21}\right],
\end{aligned}
\end{equation}
where $MLP$, namely multi-layer perception, holds a same structure with $G$ in Eq. (\ref{eq1}), as illustrated in Fig. \ref{fig:modules}. We define the different modalities as  $\mathbf{x}_{i}$ where $i\in \left\{o,t,u\right\}$ stands for the first modality, the second modality, the unlabeled samples, and the corresponding $l$-th hidden layer as $ \mathbf{z}_{i}^{(l)}$. Accordingly, the network parameters can be updated by jointly optimizing the following overall loss function.
\begin{equation}
\label{eq3}
\begin{aligned}
       L=L_{l}+L_{pl}+L_{rec}+L_{adv},
\end{aligned}
\end{equation}
where $L_{l}$ is the cross-entropy loss for labeled samples while $L_{pl}$ for pseudo-labeled samples. In addition to the two loss functions that connect the input data with labels (or pseudo-labels), we consider the reconstruction loss ($L_{rec}$) for each modality as well as unlabeled samples.
\begin{equation}
\label{eq4}
\begin{aligned}
       L_{rec}=\norm{\mathbf{x}_{o}-\mathbf{\hat{x}}_{o}}_{2}^{2}+\norm{\mathbf{x}_{t}-\mathbf{\hat{x}}_{t}}_{2}^{2}+\norm{\mathbf{x}_{u}-\mathbf{\hat{x}}_{u}}_{2}^{2},
\end{aligned}
\end{equation}
where $\mathbf{\hat{x}}_{i}$ denotes the reconstructed data of $\mathbf{x}_{i}$. For the adversarial loss ($L_{adv}$), it acts on the SA module formulated based on GANs as
\begin{equation}
\label{eq5}
\begin{aligned}
       &L_{adv}=L_{adv}^{o}+L_{adv}^{t}+L_{adv}^{u},\\
       L_{adv}^{i}=&\max_{{D}_{i}} \mathbb{E}\left[log(D_{i}(\mathbf{z}_{i}^{r}))+log(1-D_{i}(\mathbf{z}_{i}^{f}))\right],
\end{aligned}
\end{equation}
where $D_{i}$ represents the discriminator in adversarial training. Linking with Eq. (\ref{eq1}), $\mathbf{z}_{i}^{r}=(\mathbf{z}_{SA}^{1})_{i}$ and $\mathbf{z}_{i}^{f}=(\mathbf{z}_{SA}^{2})_{i}$ are a real / fake pair of data representation on the last layers of \textit{SA} module.

\subsection{Model Architecture}
The X-ModalNet starts with a \textit{feature extractor} : two convolution layers with 5$\times$5 and 3$\times$3 convolutional kernels for MSI or SAR pathway and two fully-connected layers for HSI pathway, and then passes through the \textit{SA} module with two fully-connected layers. Following it, an \textit{IL} module with two fully-connected layers is connected over the previous outputs. In the end, four fully-connected layers with an additional soft-max layer are applied to bridge the hidden layers with one-hot encoded labels. Table \ref{table:model} details the network configuration for each layer in X-ModalNet.

\begin{table*}[!t]
\centering
\caption{Network configuration in each layer of X-ModalNet. FC, Conv, and BN are abbreviations of fully connected, convolution, and batch normalization, respectively. The symbols of `$\leftrightarrow$' and `--' represent the parameter sharing and no operations, respectively. Moreover, $d_1$ and $d_2$, denote the dimensions of MSI / SAR and HSI, and $C$ is the number of class. Please note that the reconstruction happens after passing through the first block of prediction module.}
\resizebox{1\textwidth}{!}{
\begin{tabular}{c||ccccc}
\toprule[1.5pt]
& \multicolumn{5}{c}{X-ModalNet}\\
\hline \hline
Pathway & Labeled MSI / SAR ($d_1$) & & Unlabeled MSI / SAR ($d_1$) & & Labeled HSI ($d_2$)\\
\hline
\multirow{4}{*}{Feature Extractor} & $5\times 5$ Conv + BN + Dropout& \multirow{2}{*}{$\leftrightarrow$} & $5\times 5$ Conv + BN + Dropout& \multirow{2}{*}{--} & FC Encoder + BN + Dropout\\
& Tanh ($32$) & & Tanh ($32$) & & Tanh ($160$)\\
\cline{2-6}
& $3\times 3$ Conv + BN + Dropout& \multirow{2}{*}{$\leftrightarrow$} & $3\times 3$ Conv + BN + Dropout& \multirow{2}{*}{--} & FC Encoder + BN + Dropout\\
& Tanh ($64$) & & Tanh ($64$) & & Tanh ($64$)\\
\hline
\multirow{4}{*}{SA Module} & FC Encoder + BN + Dropout& \multirow{2}{*}{$\leftrightarrow$} & FC Encoder + BN  + Dropout& \multirow{2}{*}{--} & FC Encoder + BN + Dropout\\
& Tanh ($128$)& & Tanh ($128$)& & Tanh ($128$)\\
\cline{2-6}
& FC Encoder + BN + Dropout& \multirow{2}{*}{$\leftrightarrow$} & FC Encoder + BN  + Dropout& \multirow{2}{*}{--} & FC Encoder + BN + Dropout\\
& Tanh ($64$) & & Tanh ($64$) & & Tanh ($64$)\\
\hline
\multirow{4}{*}{IL Module} & FC Encoder + BN + Dropout& \multirow{2}{*}{$\leftrightarrow$} & FC Encoder + BN  + Dropout& \multirow{2}{*}{--} & FC Encoder + BN + Dropout\\
& Tanh ($64$) & & Tanh ($64$) & & Tanh ($64$)\\
\cline{2-6}
& FC Encoder + BN + Dropout& \multirow{2}{*}{$\leftrightarrow$} & FC Encoder + BN  + Dropout& \multirow{2}{*}{--} & FC Encoder + BN + Dropout\\
& Tanh ($64$) & & Tanh ($64$) & & Tanh ($64$)\\
\hline
\multirow{8}{*}{Prediction} & FC Encoder + BN + Dropout& \multirow{2}{*}{$\leftrightarrow$} & FC Encoder + BN  + Dropout& \multirow{2}{*}{--} & FC Encoder + BN + Dropout\\
& Tanh ($128$) & & Tanh ($128$) & & Tanh ($128$)\\
\cline{2-6}
& FC Encoder + BN + Dropout& & \multirow{6}{*}{FC Encoder + Softmax}& & FC Encoder + BN + Dropout\\
& Tanh ($256$) & & \multirow{6}{*}{Tanh ($C$)} & & Tanh ($256$)\\
\cline{2-2}\cline{6-6}
& FC Encoder + BN + Dropout& & & & FC Encoder + BN + Dropout\\
& Tanh ($64$) & & & & Tanh ($64$)\\
\cline{2-2}\cline{6-6}
& FC Encoder + Softmax& & & & FC Encoder + Softmax\\
& Tanh ($C$) & & & & Tanh ($C$)\\
\hline
\multirow{6}{*}{Reconstruction} & FC Encoder + BN & \multirow{2}{*}{$\leftrightarrow$} & FC Encoder + BN & \multirow{2}{*}{--} & FC Encoder + BN\\
& Tanh ($64$) & & Tanh ($64$) & & Tanh ($64$)\\
\cline{2-6}
& $3\times 3$ Conv + BN& \multirow{2}{*}{$\leftrightarrow$} & $3\times 3$ Conv + BN & \multirow{2}{*}{--} & FC Encoder + BN\\
& Tanh ($32$) & & Tanh ($32$) & & Tanh ($160$)\\
\cline{2-6}
& $5\times 5$ Conv + BN & \multirow{2}{*}{$\leftrightarrow$} & $5\times 5$ Conv + BN & \multirow{2}{*}{--} & FC Encoder + BN\\
& Sigmoid ($d_1$) & & Sigmoid ($d_1$) & & Sigmoid ($d_2$)\\
\bottomrule[1.5pt]
\end{tabular}
\label{table:model}}
\end{table*}

\section{Experiments}
\subsection{Data Description}
\begin{figure}[!t]
	  \centering
		\subfigure[HSI-MSI datasets]{
			\includegraphics[width=0.85\textwidth]{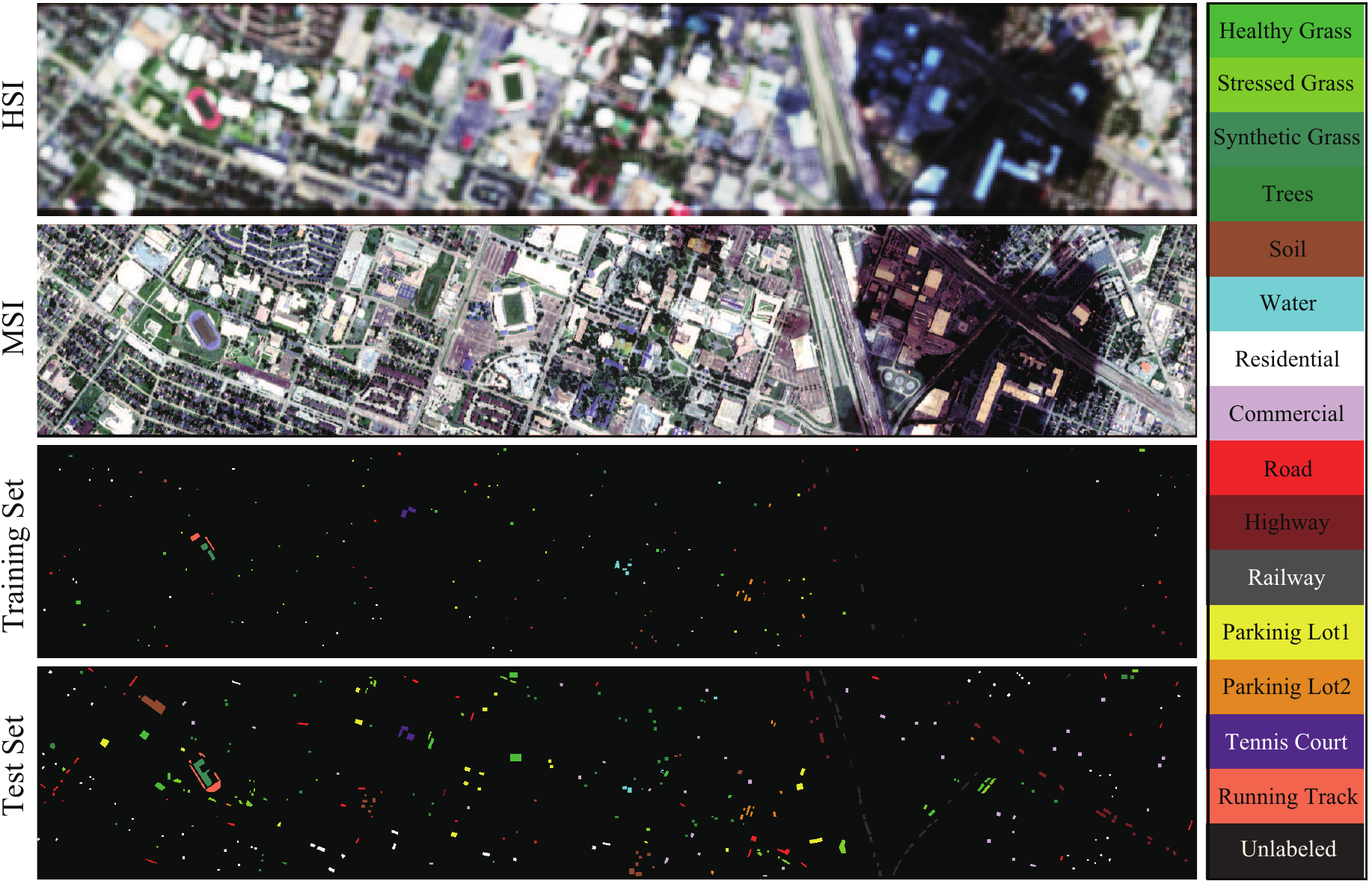}
    		}
        \subfigure[HSI-SAR datasets]{
			\includegraphics[width=0.65\textwidth]{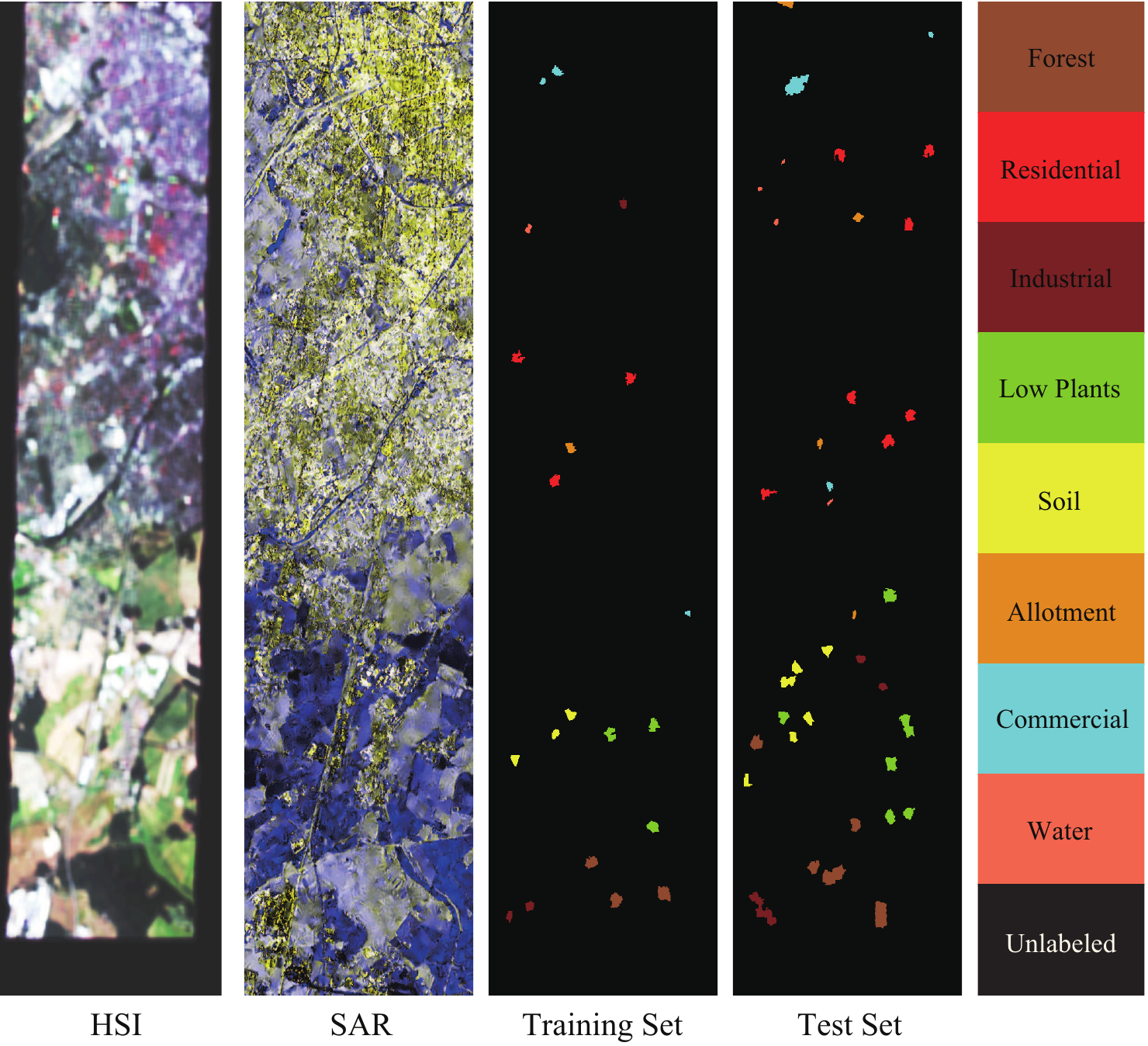}
		}
        \caption{Exemplary datasets for HSI-MSI and HSI-SAR: false-color images and corresponding training and test labels.}
\label{fig:Faslecolor}
\end{figure}

We evaluate the performance of the X-ModalNet on two different datasets. Fig. \ref{fig:Faslecolor} shows the false-color images for both datasets as well as the corresponding training and test ground truth maps, while scene categories and the number of training and test samples are detailed in Table \ref{table:trainingtest}. {\bf There are two things particularly noteworthy in our CML' s setting:} 1) \textit{vector (or patch)-based} input due to the sparse groundtruth maps; 2) we assume that \textit{the HSI is present only in the process of training and it is absent in the test phase.}

\subsubsection{Homogeneous HSI-MSI Dataset}
The HSI scene that has been widely used in many works \citep{hong2017learning,hong2020invariant} consists of $349\times1905$ pixels with 144 spectral bands in the wavelength range from 380 $nm$ to 1050 $nm$ at a ground sampling distance (GSD) of 10 $m$ (low spatial-resolution), while the aligned MSI with the dimensions of $349\times1905\times8$ is obtained at a GSD of 2.5 $m$ (high spatial-resolution).

\textit{Spectral simulation} is performed to generate the low-spectral resolution MSI by degrading the reference HSI in the spectral domain using the MS spectral response functions of Sentinel-2 as filters. Using this, the MSI consists of $349\times1905$ pixels with eight spectral bands at a GSD of 2.5 $m$.

\textit{Spatial simulation} is performed to generate the low-spatial resolution HSI by degrading the reference HSI in the spatial domain using an isotropic Gaussian point spread function, thus yielding the HSI with the dimensions of $349\times1905\times144$ at a GSD of 10 $m$ by upsampling to the MSI's size.

\begin{table}[!t]
\centering
\caption{The number of training and test samples on two datasets.}
\resizebox{0.75\textwidth}{!}{
\begin{tabular}{l||ccc||ccc}
\toprule[1.5pt]
Dataset & \multicolumn{3}{c||}{HSI-MSI}&\multicolumn{3}{c}{HSI-SAR}\\
\hline \hline No.&Class&Training&Test&Class&Training&Test\\
\hline \hline 1&Healthy Grass&537&699&Forest&1437&3249\\
 2&Stressed Grass&61&1154&Residential&961&2373\\
 3&Synthetic Grass&340&357&Industrial&623&1510\\
 4&Tree&209&1035&Low Plants&1098&2681\\
 5&Soil&74&1168&Soil&728&1817\\
 6&Water&22&303&Allotment&260&747\\
 7&Residential&52&1203&Commercial&451&1313\\
 8&Commercial&320&924&Water&144&256\\
 9&Road&76&1149&--&--&--\\
 10&Highway&279&948&--&--&--\\
 11&Railway&33&1185&--&--&--\\
 12&Parking Lot1&329&904&--&--&--\\
 13&Parking Lot2&20&449&--&--&--\\
 14&Tennis Court&266&162&--&--&--\\
 15&Running Track&279&381&--&--&--\\
 \hline \hline & Total & 2832 & 12197 & Total & 5702 & 13946 \\
\bottomrule[1.5pt]
\end{tabular}
}
\label{table:trainingtest}
\end{table}

\subsubsection{Heterogeneous HSI-SAR Dataset}
The EnMap benchmark HSI covering the Berlin urban area is freely available from the website\footnote{{\url{ http://doi.org/10.5880/enmap.2016.002}}}. This image consists of $797\times220$ pixels with a GSD of 30 $m$, and 244 spectral bands ranging from 400 $nm$ to 2500 $nm$. According to the geographic coordinates, we downloaded the same scene of SAR image from the Sentinel-1 satellite, with the size of $1723\times476$ pixels at a GSD of 13 $m$ and four polarimetric bands \citep{yamaguchi2005four}. The used SAR image is dual-polarimetric SAR data collected by interferometric wide swath mode. It is organized as a commonly used four-component PolSAR covariance matrix (four bands) \citep{yamaguchi2005four}. Note that we upsample the HSI to the same size with the SAR image by the nearest-neighbor interpolation.

\subsection{Implementation Details}
Our approach is implemented on the Tensorflow framework \citep{abadi2016tensorflow}. The network configuration, to our knowledge, always plays a critical role in a practical DL system. The model is trained on the training set, and the hyper-parameters are determined using a grid search on the validation set\footnote{Ten replications are conducted to randomly split the original training set into the new training and validation sets with the percentage of 8:2 to determine the network's hyperparameters.}. In the training phase, we adopt the Adam optimizer with the ``poly'' learning rate policy \citep{chen2018deeplab}. The current learning rate can be updated by multiplying the base one with $(1-\frac{iter}{maxIter})^{power}$, where the base learning rate and power are set to 0.0005 and 0.98, respectively. We use the DAE to pretrain the subnetworks for each modality to greatly reduce the training time of the model and find a better local optimum easier. Also, the momentum is set to 0.9.

To facilitate network training and reduce overfitting, BN and dropout techniques are orderly used for all DL-based methods prior to the activation functions. The model training ends up with 150 epochs for the heterogeneous HSI-MSI dataset and 200 epochs for the heterogeneous HSI-SAR dataset with a minibatch size of 300. Both labeled and unlabeled samples in SAR or MSI share the same network parameters in the process of model optimization.

In the experiments, we found that when the unlabeled samples, from neither training nor test sets, are selected at an approximated scale with the test set, the final classification results are similar to that directly using test set. We have to admit, however, that the full use of unlabeled samples enable further improvement in classification performance, but we have to make a trade-off between the limited performance improvement and exponentially increasing cost in data storage, transmission, and computation. Moreover, we expect to see the performance gain when using these proposed modules, thereby demonstrating their effectiveness and superiority. As a result, we, for simplicity, select the test set as the unlabeled set for all semi-supervised compared methods for a fair comparison.

Furthermore, two commonly used indices: \textit{Pixel-wise Accuracy} (Pixel Acc.) and \textit{mean Intersection over Union} (mIoU) are calculated to quantitatively evaluate the parsing performance by collecting all pixel-wise predictions of the test set. Due to random initialization, both metrics show the average accuracy and the variation of the results out of 10 runs.

\subsection{Comparison with State-of-the-art}
Several state-of-the-art baselines closely related to our task (CML) are selected for comparison; they are

1) \textbf{Baseline:} We train a linear SVM classifier directly using original pixel-based MSI or SAR features. Note that the hyperparameters in SVM are determined by 10-fold cross-validation on the training set.

2) \textbf{Canonical Correlation Analysis (CCA)} \citep{hardoon2004canonical}: We learn a shared latent subspace from two modalities on the training set, and project the test samples from any one of the two modalities into the subspace. This is a typical cross-modal feature learning. Finally, the learned features are fed into a linear SVM. We used the code from the website\footnote{\url{https://github.com/CommonClimate/CCA}}.

3) \textbf{Unimodal DAE} \citep{chen2014deep}: This is a classical deep autoencoder. We train a DAE on the target-modality (MSI or SAR) in a unsupervised way, and finely tune it using labels. The hidden representation of the encoder layer is used for final classification. The code we used is from the website\footnote{\url{http://deeplearning.stanford.edu/wiki/index.php/UFLDL$\_$Tutorial}}.

4) \textbf{Bimodal DAE} \citep{ngiam2011multimodal}: As a DL's pioneer to multi-modal application, it learns a joint feature representation over the encoder layers generated by AEs for each modality.

5) \textbf{Bimodal SDAE} \citep{silberer2017visually}: This is a semi-supervised version for Bimodal DAE by considering the reconstruct loss of all unlabeled samples for each modality and adding an additional soft-max layer over the encoder layer for those limited labeled data.

6) \textbf{MDL-CW} \citep{rastegar2016mdl}: A end-to-end multimodal network is trained with cross weights acted on the two-stream subnetworks for more effective information blending.

7) \textbf{Corr-AE} \citep{feng2014cross}: A coupled AEs are first used to learn a shared high-level feature representation by enforcing similarity constraint between the encoder layers of two modalities. The learned features are then fed into a classifier.

8) \textbf{CorrNet} \citep{chandar2016correlational}: Similar to Corr-AE, AE is responsible for extracting features of each modality, while CCA serves as a link with the features by maximizing their correlations. The code is available from the website\footnote{\url{https://github.com/apsarath/CorrNet}}.

\begin{table*}[!t]
\centering
\caption{Quantitative performance comparison with baseline models on the HSI-MSI dataset. The best one is shown in bold.}
\resizebox{0.6\textwidth}{!}{
\begin{tabular}{lcc}
\toprule[1.5pt]
Methods & Pixel Acc. (\%) & mIoU (\%)\\
\hline\hline
Baseline & 70.51 & 57.84\\
CCA \citep{hardoon2004canonical} & 73.01 & 64.72\\
Unimodal DAE \citep{chen2014deep} & 72.85 $\pm$ 1.2 & 62.75 $\pm$ 0.3\\
Bimodal DAE \citep{ngiam2011multimodal} & 75.43 $\pm$ 0.6 & 67.67 $\pm$ 0.1\\
Bimodal SDAE \citep{silberer2017visually} & 79.51 $\pm$ 1.7 & 69.62 $\pm$ 0.3\\
MDL-CW \citep{rastegar2016mdl} & 83.27 $\pm$ 1.0 & 74.60 $\pm$ 0.3\\
Corr-AE \citep{feng2014cross} & 80.49 $\pm$ 1.2 & 70.85 $\pm$ 0.2\\
CorrNet \citep{chandar2016correlational} & 82.66 $\pm$ 0.8 & 73.48 $\pm$ 0.2\\
X-ModalNet & \bf{88.23 $\pm$ 0.7} & \bf{80.31 $\pm$ 0.2}\\
\bottomrule[1.5pt]
\end{tabular}
}
\label{table:Performance_Houston}
\end{table*}

\begin{table*}[!t]
\centering
\caption{Quantitative performance comparison with baseline models on the HSI-SAR datasets. The best one is shown in bold.}
\resizebox{0.6\textwidth}{!}{
\begin{tabular}{lcc}
\toprule[1.5pt]
Methods & Pixel Acc. (\%) & mIoU (\%)\\
\hline\hline
Baseline & 43.91 & 18.70\\
CCA  \citep{hardoon2004canonical} & 36.66 & 12.04\\
Unimodal DAE \citep{chen2014deep} & 51.51 $\pm$ 0.5 & 29.32 $\pm$ 0.2\\
Bimodal DAE \citep{ngiam2011multimodal} & 56.04 $\pm$ 0.5 & 34.13 $\pm$ 0.2\\
Bimodal SDAE \citep{silberer2017visually} & 59.27 $\pm$ 0.6 & 37.78 $\pm$ 0.2\\
MDL-CW \citep{rastegar2016mdl} & 62.51 $\pm$ 0.8 & 42.15 $\pm$ 0.1\\
Corr-AE \citep{feng2014cross} & 60.59 $\pm$ 0.5 & 39.12 $\pm$ 0.3\\
CorrNet \citep{chandar2016correlational} & 64.65 $\pm$ 0.7 & 44.25 $\pm$ 0.3\\
X-ModalNet & \bf{71.38 $\pm$ 1.0} & \bf{54.02 $\pm$ 0.3}\\
\bottomrule[1.5pt]
\end{tabular}
}
\label{table:Performance_Berlin}
\end{table*}

\begin{figure*}[!t]
\centering
\includegraphics[width=1\textwidth]{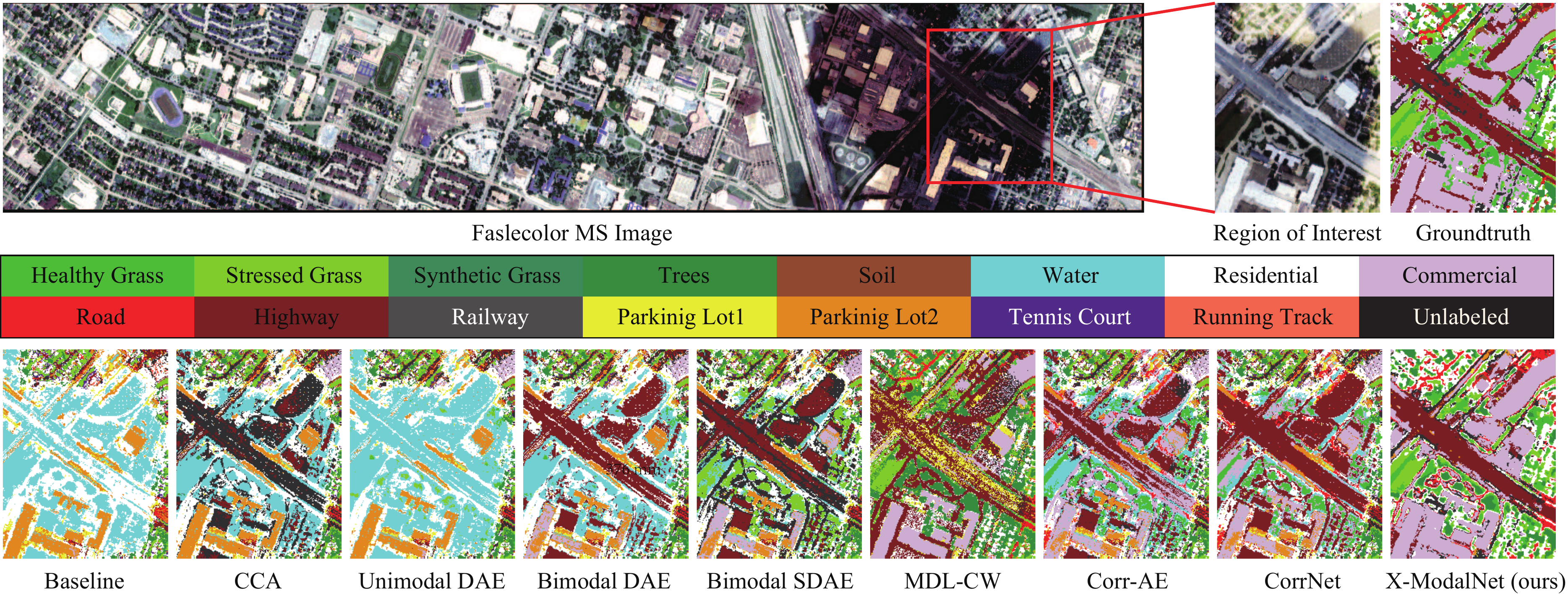}
\caption{Classification maps of ROI on HSI-MSI datasets. The ground truth in this highlighted area is manually labelled.}
\label{fig:CM_MSI}
\end{figure*}

\begin{figure*}[!t]
\centering
\includegraphics[width=1\textwidth]{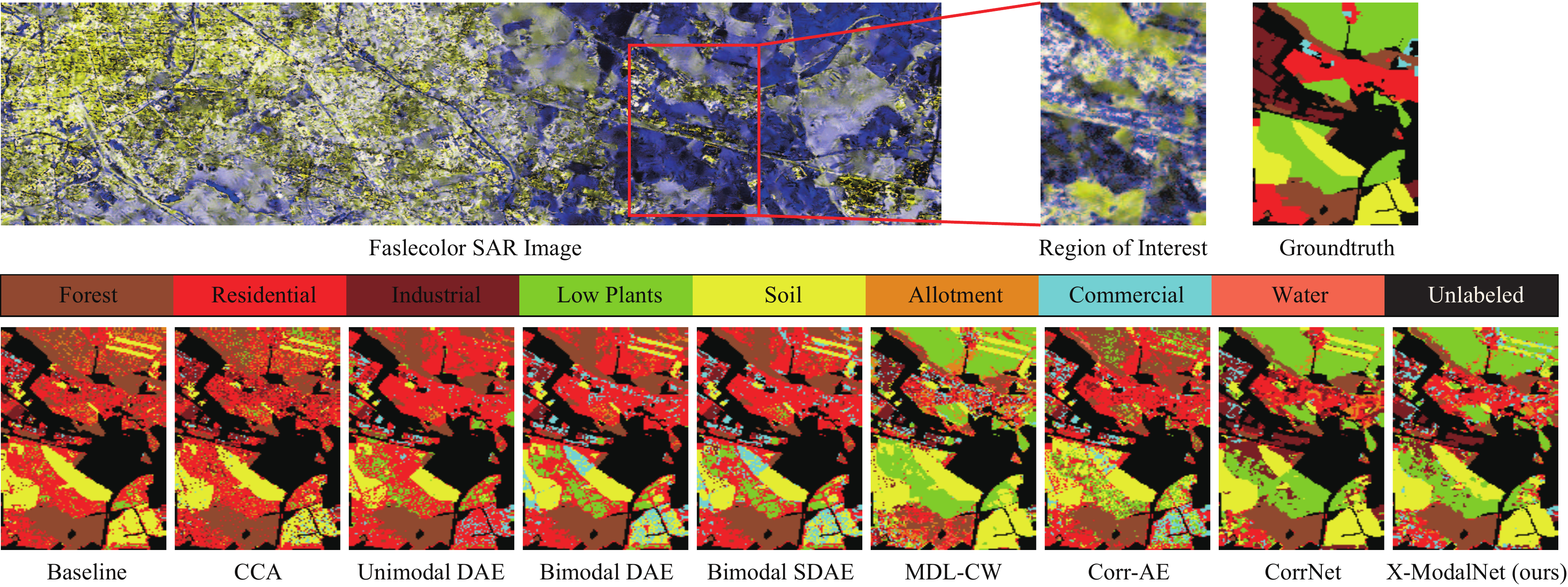}
\caption{Classification maps of ROI on HSI-SAR datasets. The OpenStreetMap \protect \citep{haklay2008openstreetmap} is used as the ground truth generator for this area.}
\label{fig:CM_SAR}
\end{figure*}

\subsubsection{Results on the Homogeneous Datasets}
Table \ref{table:Performance_Houston} shows the quantitative performance comparison in terms of Pixel Acc. and mIoU. Limited by the feature diversity, the baseline yields a poor classification performance, while there is a performance improvement (about $2\%$) in the unimodal DAE due to the powerful learning ability of DL-based techniques. For the homogeneous HSI-MSI correspondences, the linearized CCA is more likely to catch the shared features and obtains the better classification results. The features can be better fused over the hidden representations of two modalities. Therefore, the bimodal DAE improves the performance by $2\%$ on the basis of CCA's. The accuracy of bimodal SDAE can further increase to around $79\%$, since it aims at training an end-to-end multimodal network to generate more discriminative features. Different from previous strategies, Corr-AE and CorrNet couple two subnetworks by enforcing the structural measurement on hidden layers, such as Euclidean similarity and correlation, which allows a more effective pixel-wise classification. The MDL-CW with learning cross weights can facilitate the multimodal information fusion, thus achieving better classification results than Corr-AE and CorrNet. As expected, X-ModalNet outperforms these state-of-the-art methods, demonstrating its superiority and effectiveness with a large improvement of at least 6\% Pixel Acc. and mIoU over CorrNet (the second best method).

\subsubsection{Results on the Heterogeneous Datasets}
Similar to the former dataset, we evaluate the performance for the Heterogeneous HSI-SAR scene quantitatively. Two assessment indices (Pixel Acc. and mIoU) for different algorithms are summarized in Table \ref{table:Performance_Berlin}. There is a basically consistent trend in performance improvement of different algorithms. That is, the performance of X-ModalNet is significantly superior to that of others, and the methods with the hyperspectral information perform better than those without one, such as Baseline and Unimodal DAE. It is worth noting that the proposed X-ModalNet brings increments of about 9\% Pixel Acc. and 10\% mIoU on the basis of CorrNet. Moreover, the CCA fails to linearly represent the heterogeneous data, leading to a worse parsing result and even lower than the baseline. Additionally, the gap (or heterogeneity) between SAR and optical data can be effectively reduced by mutually learning weights. This might explain the case that the MDL-CW observably exceeds most compared methods without such interactive module (nearly $20\%$ over baseline), e.g., Bimodal DAE and its semi-supervised version (Bimodal SDAE) as well as CorrNet.

\subsection{Visual Comparison}
Apart from quantitative assessment, we also make a visual comparison by highlighting a salient region overshadowed by the cloud on the Houston2013 datasets. As shown in Fig. \ref{fig:CM_MSI}, our method is capable of identifying various materials more effectively, particularly for the material \textit{Commercial} in the upper-right of the predicted maps. Besides, a trend can be figured out, that is, the methods with the input of multi-modalities achieve more smooth parsing results compared to those with the input of single modalities.

Similarly, we visually show the classification maps of those comparative algorithms in a region of interest in the EnMap datasets, as shown in Fig. \ref{fig:CM_SAR}. We can see that our X-ModalNet shows a more competitive and realistic parsing result, especially in classifying \textit{Soil} and \textit{Plants}, which is more approaching to the real scene.

\begin{table*}[!t]
\centering
\caption{Ablation analysis of the X-ModalNet with a combination of different modules in term of Pixel Acc. on two datasets. Moveover, importance analysis in the presence and absence of BN and dropout operations is discussed as well.}
\resizebox{0.75\textwidth}{!}{
\begin{tabular}{lccccccc}
\toprule[1.5pt]
\multirow{2}{*}{Methods} &\multirow{2}{*}{BN} &\multirow{2}{*}{Dropout} &\multirow{2}{*}{IL} &\multirow{2}{*}{LP} &\multirow{2}{*}{SA} & \multicolumn{2}{c}{Pixel Acc. (\%)}\\
\cline{7-8}& & & & & & HSI-MSI & HSI-SAR\\
\hline\hline
X-ModalNet & $\checkmark$ & $\checkmark$ & $\times$ & $\times$ & $\times$ & 83.14 $\pm$ 0.9 & 64.44 $\pm$ 1.1\\
X-ModalNet & $\checkmark$ & $\checkmark$ & $\checkmark$ & $\times$ & $\times$ & 85.07 $\pm$ 0.8 & 68.73 $\pm$ 0.8\\
X-ModalNet & $\checkmark$ & $\checkmark$ & $\checkmark$ & $\checkmark$ & $\times$ & 86.58 $\pm$ 1.0 & 70.19 $\pm$ 0.8\\
X-ModalNet& $\checkmark$ & $\checkmark$ & $\checkmark$ & $\checkmark$ & $\checkmark$ & 88.23 $\pm$ 0.7 & 71.38 $\pm$ 1.0\\
X-ModalNet& $\times$ & $\times$ & $\checkmark$ & $\checkmark$ & $\checkmark$ & 80.33 $\pm$ 0.5 & 62.47 $\pm$ 0.7\\
X-ModalNet& $\times$ & $\checkmark$ & $\checkmark$ & $\checkmark$ & $\checkmark$ & 81.94 $\pm$ 0.6 & 64.10 $\pm$ 0.9\\
X-ModalNet& $\checkmark$ & $\times$ & $\checkmark$ & $\checkmark$ & $\checkmark$ & 85.10 $\pm$ 0.6 & 67.34 $\pm$ 0.8\\
\bottomrule[1.5pt]
\end{tabular}
}
\label{table:Ablation}
\end{table*}

\begin{figure*}[!t]
	  \centering
		\subfigure[HSI-MSI datasets]{
			\includegraphics[width=0.8\textwidth]{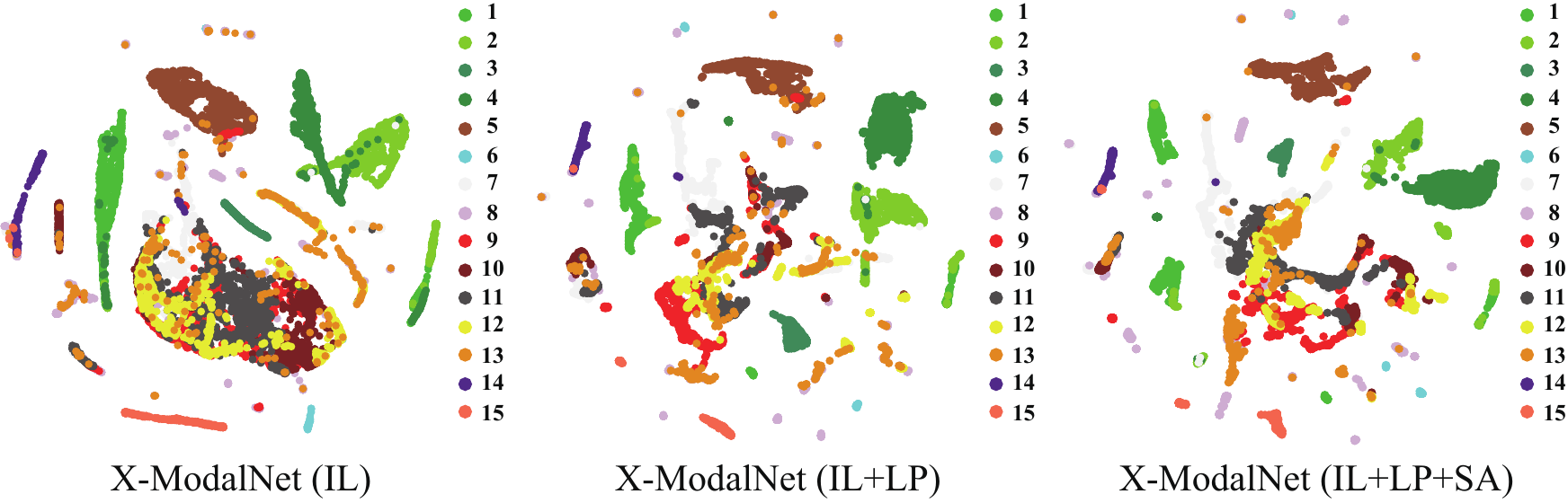}
    		}
        \subfigure[HSI-SAR datasets]{
			\includegraphics[width=0.8\textwidth]{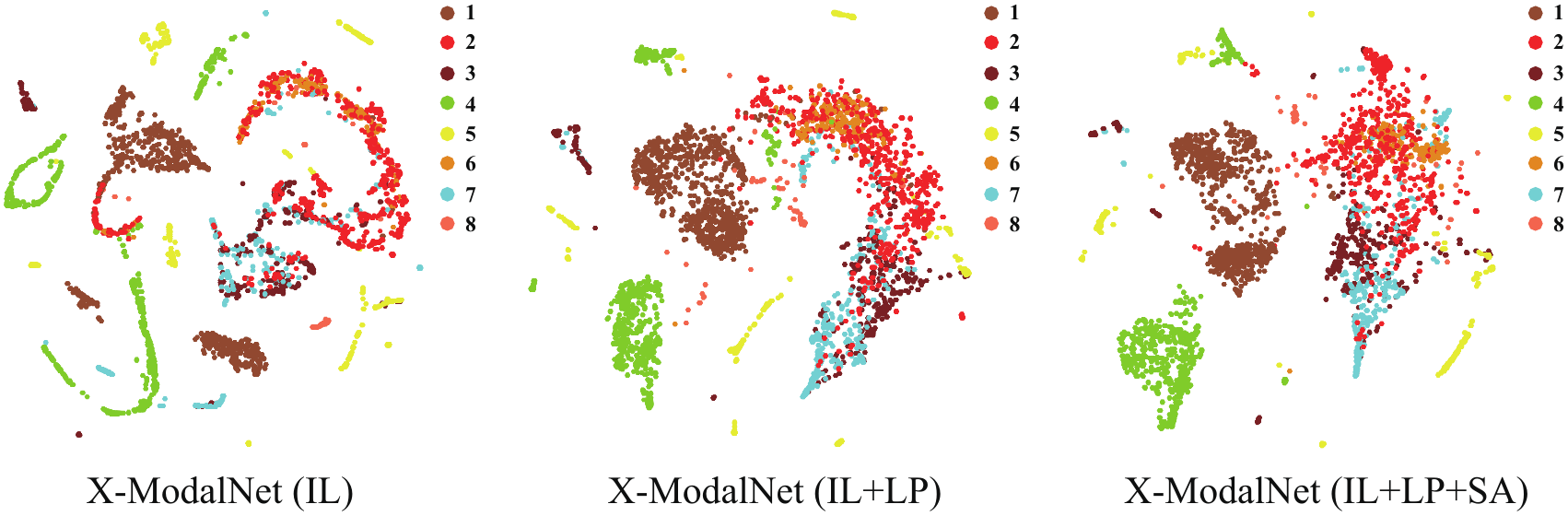}
		}
        \caption{t-SNE visualization of the learned multimodal features in the latent space using X-ModalNet with different modules on the two different datasets.}
\label{fig:visualization}
\end{figure*}

\subsection{Ablation Studies}
We analyze the performance gain of X-ModalNet by step-wise adding the different components (or modules). Table \ref{table:Ablation} lists a progressive performance improvement by gradually embedding different modules, while Fig. \ref{fig:visualization} correspondingly visualizes the learned features in the latent space (top encoder layer). It is clear to observe that successively adding each component into the X-ModalNet is conducive to a more discriminative feature generation.

We also investigate the importance of dropout and BN techniques in avoiding overfitting and improving network performance. As can be seen in Table \ref{table:Ablation}, turning off the dropout would hinder X-ModalNet from generalizing well, yielding a performance degradation. What is worse is that the classification accuracy without BN reduces sharply. This could result from low-efficiency gradient propagation, thereby hurting the learning ability of the network. Moreover, we can observe from Table \ref{table:Ablation} that the classification performance without any proposed modules is limited, only yielding about $83.14\%$ and $64.44\%$ Pixel Acc. on the two datasets. It is worth noting that the results achieve an obvious improvement (around $2\%\sim3\%$) after plugging the IL module. By introducing the semi-supervised mechanism, our LP module can bring increments of  $1.5\%$ and $2\%$ Pixel Acc. on the basis of only using the IL module for HSI-MSI and HSI-SAR, respectively. Remarkably, when adding the SA module over the IL and LP modules in networks, our X-ModalNet behaves superiorly and obtains a further dramatic improvement in classification accuracies. These, to a great extent, demonstrate the effectiveness and superiority of several proposed modules as well as their positive effects on the classification performance.

\begin{figure*}[!t]
	  \centering
		\subfigure[HSI-MSI datasets]{
			\includegraphics[width=0.42\textwidth]{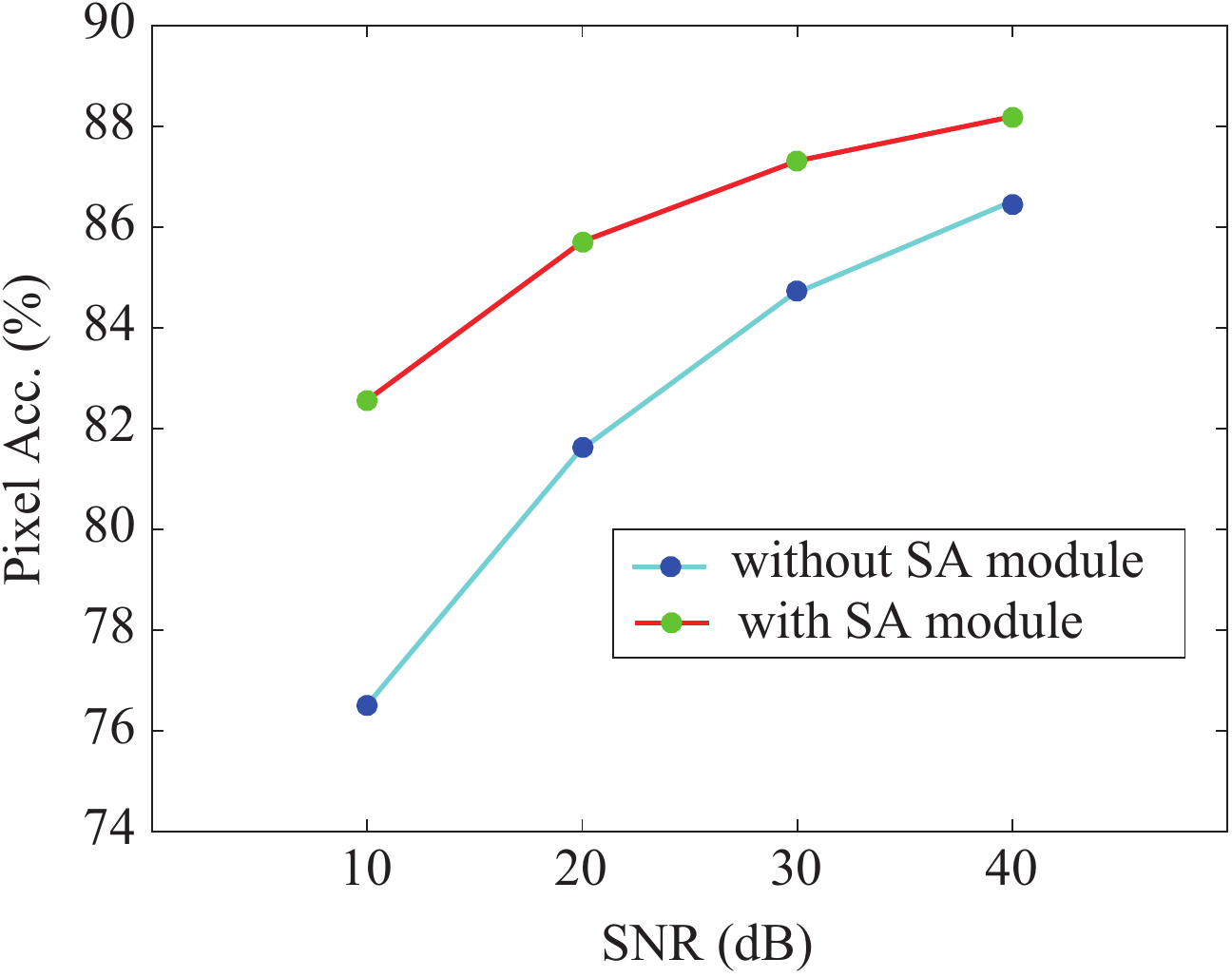}
    		}\qquad
        \subfigure[HSI-SAR datasets]{
			\includegraphics[width=0.42\textwidth]{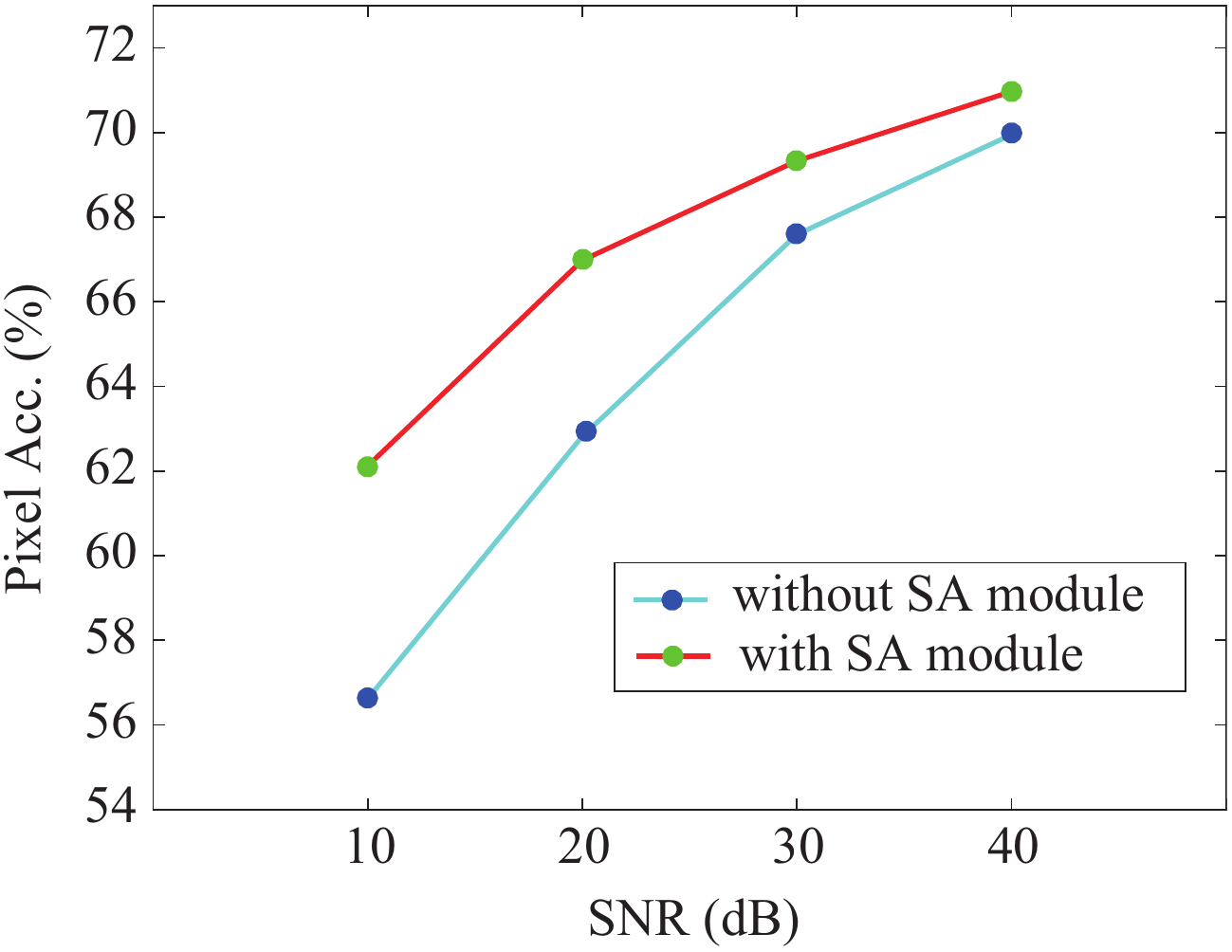}
		}
        \caption{Resistance analysis to noise attack using the proposed X-ModalNet with and without \textit{SA} module on the two datasets.}
\label{fig:noise_attack}
\end{figure*}

\subsection{Robustness to Noises}
Neural networks have shown their vulnerability to adversarial samples generated by slight perturbation, e.g., imperceptible noises. To study the effectiveness of our SA module against noise or perturbation attack, we simulate the corrupted input by adding Gaussian white noises with different signal-to-noise-ratios (SNRs) ranging from 10 dB to 40 dB at a 10 dB interval. Fig. \ref{fig:noise_attack} shows a quantitative comparison in term of Pixel Acc. before and after triggering the \textit{SA} module.

\section{Conclusion}
In this paper, we investigate the cross-modal classification task by utilizing multimodal satellite or aerial images (RS data). In reality, the HSI is only able to be collected in a locally small area due to the limitations of the imaging system, while MSI and SAR are openly available on a global scale. This motivate us to learn to transfer the HSI knowledge into large-scale MSI or SAR by training the model on both modalities and predict only on one modality. To address the CML's issue in RS, we propose a novel DL-based model X-ModalNet, with two well-designed components (IL and SA modules) to effectively learn a more discriminative feature representation and robustly resist the noise attack, respectively, and with an iteratively updating LP mechanism for further improving the network performance. In the future work, we would like to introduce the physical mechanism of spectral imaging into the network learning for earth observation tasks.

\section*{Acknowledgement}
This work is jointly supported by the German Research Foundation (DFG) under grant ZH 498/7-2, by the European Research Council (ERC) under the European Union's Horizon 2020 research and innovation programme (grant agreement No. [ERC-2016-StG-714087], Acronym: \textit{So2Sat}), by the Helmholtz Association
through the framework of Helmholtz Artificial Intelligence (HAICU) - Local Unit ``Munich Unit @Aeronautics, Space and Transport (MASTr)'' and Helmholtz Excellent Professorship ``Data Science in Earth Observation - Big Data Fusion for Urban Research'', by the German Federal Ministry of Education and Research (BMBF) in the framework of the international AI future lab "AI4EO -- Artificial Intelligence for Earth Observation: Reasoning, Uncertainties, Ethics and Beyond", by the National Natural Science Foundation of China (NSFC) under grant contracts No.41820104006, as well as by the AXA Research Fund. This work of N. Yokoya is also supported by the Japan Society for the Promotion of Science (KAKENHI 18K18067).

\bibliographystyle{elsarticle-harv}
\bibliography{mybibfile}

\begin{thebibliography}{90}
\expandafter\ifx\csname natexlab\endcsname\relax\def\natexlab#1{#1}\fi
\expandafter\ifx\csname url\endcsname\relax
  \def\url#1{\texttt{#1}}\fi
\expandafter\ifx\csname urlprefix\endcsname\relax\def\urlprefix{URL }\fi

\bibitem[{Abadi et~al.(2016)Abadi, Barham, Chen, Z.Chen, Davis, Dean, Devin,
  Ghemawat, Irving, and Isard}]{abadi2016tensorflow}
Abadi, M., Barham, P., Chen, J., Z.Chen, Davis, A., Dean, J., Devin, M.,
  Ghemawat, S., Irving, G., Isard, M., 2016. Tensorflow: a system for
  large-scale machine learning. In: OSDI. Vol.~16. pp. 265--283.

\bibitem[{Audebert et~al.(2016)Audebert, Saux, and
  Lef{\`e}vre}]{audebert2016semantic}
Audebert, N., Saux, B.~L., Lef{\`e}vre, S., 2016. Semantic segmentation of
  earth observation data using multimodal and multi-scale deep networks. In:
  Proc. ACCV. Springer, pp. 180--196.

\bibitem[{Audebert et~al.(2017)Audebert, Saux, and
  Lef{\`e}vre}]{audebert2017joint}
Audebert, N., Saux, B.~L., Lef{\`e}vre, S., 2017. Joint learning from earth
  observation and openstreetmap data to get faster better semantic maps. In:
  Proc. CVPR Workshop. IEEE, pp. 1552--1560.

\bibitem[{Audebert et~al.(2018)Audebert, Saux, and
  Lef{\`e}vre}]{audebert2018beyond}
Audebert, N., Saux, B.~L., Lef{\`e}vre, S., 2018. Beyond rgb: Very high
  resolution urban remote sensing with multimodal deep networks. ISPRS J.
  Photogramm. Remote Sens. 140, 20--32.

\bibitem[{Badrinarayanan et~al.(2017)Badrinarayanan, Kendall, and
  Cipolla}]{badrinarayanan2017segnet}
Badrinarayanan, V., Kendall, A., Cipolla, R., 2017. Segnet: A deep
  convolutional encoder-decoder architecture for image segmentation. IEEE
  Trans. Pattern Anal. Mach. Intell.~(12), 2481--2495.

\bibitem[{Baltru{\v{s}}aitis et~al.(2018)Baltru{\v{s}}aitis, Ahuja, and
  Morency}]{baltruvsaitis2018multimodal}
Baltru{\v{s}}aitis, T., Ahuja, C., Morency, L., 2018. Multimodal machine
  learning: A survey and taxonomy. IEEE Trans. Pattern Anal. Mach. Intell.

\bibitem[{Biggio and Roli(2018)}]{biggio2018wild}
Biggio, B., Roli, F., 2018. Wild patterns: Ten years after the rise of
  adversarial machine learning. Pattern Recognit.

\bibitem[{Cangea et~al.(2017)Cangea, Veli{\v{c}}kovi{\'c}, and
  Li{\`o}}]{cangea2017xflow}
Cangea, C., Veli{\v{c}}kovi{\'c}, P., Li{\`o}, P., 2017. Xflow: 1d-2d
  cross-modal deep neural networks for audiovisual classification. arXiv
  preprint arXiv:1709.00572.

\bibitem[{Cao et~al.(2020{\natexlab{a}})Cao, Yao, Fu, Bi, and Hong}]{cao2020an}
Cao, X., Yao, J., Fu, X., Bi, H., Hong, D., 2020{\natexlab{a}}. An enhanced
  3-dimensional discrete wavelet transform for hyperspectral image
  classification. IEEE Geosci. and Remote Sens. Lett.10.1109/LGRS.2020.2990407.

\bibitem[{Cao et~al.(2020{\natexlab{b}})Cao, Yao, Xu, and
  Meng}]{cao2020hyperspectral}
Cao, X., Yao, J., Xu, Z., Meng, D., 2020{\natexlab{b}}. Hyperspectral image
  classification with convolutional neural network and active learning. IEEE
  Trans. Geosci. Remote Sens.DOI:10.1109/TGRS.2020.2964627.

\bibitem[{Chandar et~al.(2016)Chandar, Khapra, Larochelle, and
  Ravindran}]{chandar2016correlational}
Chandar, S., Khapra, M., Larochelle, H., Ravindran, B., 2016. Correlational
  neural networks. Neural computation 28~(2), 257--285.

\bibitem[{Chen et~al.(2018)Chen, Papandreou, Kokkinos, Murphy, and
  Yuille}]{chen2018deeplab}
Chen, L., Papandreou, G., Kokkinos, I., Murphy, K., Yuille, A.~L., 2018.
  Deeplab: Semantic image segmentation with deep convolutional nets, atrous
  convolution, and fully connected crfs. IEEE Trans. Pattern Anal. Mach.
  Intell. 40~(4), 834--848.

\bibitem[{Chen et~al.(2016)Chen, Jiang, Li, Jia, and Ghamisi}]{chen2016deep}
Chen, Y., Jiang, H., Li, C., Jia, X., Ghamisi, P., 2016. Deep feature
  extraction and classification of hyperspectral images based on convolutional
  neural networks. IEEE Trans. Geosci. Remote Sens. 54~(10), 6232--6251.

\bibitem[{Chen et~al.(2014)Chen, Lin, Zhao, Wang, and Gu}]{chen2014deep}
Chen, Y., Lin, Z., Zhao, X., Wang, G., Gu, Y., 2014. Deep learning-based
  classification of hyperspectral data. IEEE J. Sel. Topics Appl. Earth Observ.
  Remote Sens. 7~(6), 2094--2107.

\bibitem[{Demir et~al.(2018)Demir, Koperski, Lindenbaum, Pang, Huang, Basu,
  Hughes, Tuia, and Raskar}]{demir2018deepglobe}
Demir, I., Koperski, K., Lindenbaum, D., Pang, G., Huang, J., Basu, S., Hughes,
  F., Tuia, D., Raskar, R., 2018. Deepglobe 2018: A challenge to parse the
  earth through satellite images. Proc. CVPR Workshop.

\bibitem[{Donahue et~al.(2016)Donahue, Kr{\"a}henb{\"u}hl, and
  Darrell}]{donahue2016adversarial}
Donahue, J., Kr{\"a}henb{\"u}hl, P., Darrell, T., 2016. Adversarial feature
  learning. arXiv preprint arXiv:1605.09782.

\bibitem[{Feng et~al.(2014)Feng, Wang, and Li}]{feng2014cross}
Feng, F., Wang, X., Li, R., 2014. Cross-modal retrieval with correspondence
  autoencoder. In: Proc. ACMMM. ACM, pp. 7--16.

\bibitem[{Frome et~al.(2013)Frome, Shlens, s.~Bengio, Dean, and
  Mikolov}]{frome2013devise}
Frome, A., Shlens, G. S. C.~J., s.~Bengio, Dean, J., Mikolov, T., 2013. Devise:
  A deep visual-semantic embedding model. In: Proc. NIPS. pp. 2121--2129.

\bibitem[{Gao et~al.(2017{\natexlab{a}})Gao, Yao, Li, Zhuang, Zhang, and
  Bioucas-Dias}]{gao2017new}
Gao, L., Yao, D., Li, Q., Zhuang, L., Zhang, B., Bioucas-Dias, J.,
  2017{\natexlab{a}}. A new low-rank representation based hyperspectral image
  denoising method for mineral mapping. Remote Sens. 9~(11), 1145.

\bibitem[{Gao et~al.(2017{\natexlab{b}})Gao, Zhao, Jia, Liao, and
  Zhang}]{gao2017optimized}
Gao, L., Zhao, B., Jia, X., Liao, W., Zhang, B., 2017{\natexlab{b}}. Optimized
  kernel minimum noise fraction transformation for hyperspectral image
  classification. Remote Sens. 9~(6), 548.

\bibitem[{Ghosh et~al.(2018)Ghosh, Ehrlich, Shah, Davis, and
  Chellappa}]{ghosh2018stacked}
Ghosh, A., Ehrlich, M., Shah, S., Davis, L., Chellappa, R., 2018. Stacked
  u-nets for ground material segmentation in remote sensing imagery. In: Proc.
  CVPR Workshop. pp. 257--261.

\bibitem[{G{\'o}mez-Chova et~al.(2015)G{\'o}mez-Chova, Tuia, Moser, and
  Camps-Valls}]{gomez2015multimodal}
G{\'o}mez-Chova, L., Tuia, D., Moser, G., Camps-Valls, G., 2015. Multimodal
  classification of remote sensing images: A review and future directions.
  Proc. IEEE 103~(9), 1560--1584.

\bibitem[{Goodfellow et~al.(2014{\natexlab{a}})Goodfellow, Pouget-Abadie,
  Mirza, Xu, Warde-Farley, Ozair, Courville, and
  Bengio}]{goodfellow2014generative}
Goodfellow, I., Pouget-Abadie, J., Mirza, M., Xu, B., Warde-Farley, D., Ozair,
  S., Courville, A., Bengio, Y., 2014{\natexlab{a}}. Generative adversarial
  nets. In: Proc. NIPS. pp. 2672--2680.

\bibitem[{Goodfellow et~al.(2014{\natexlab{b}})Goodfellow, Shlens, and
  Szegedy}]{goodfellow2014explaining}
Goodfellow, I., Shlens, J., Szegedy, C., 2014{\natexlab{b}}. Explaining and
  harnessing adversarial examples. arXiv:1412.6572.

\bibitem[{Haklay and Weber(2008)}]{haklay2008openstreetmap}
Haklay, M., Weber, P., 2008. Openstreetmap: User-generated street maps. IEEE
  Pervasive Computing 7~(4), 12--18.

\bibitem[{Han et~al.(2018)Han, Huang, Li, Li, Yang, and Gong}]{han2018edge}
Han, X., Huang, X., Li, J., Li, Y., Yang, M., Gong, J., 2018. The
  edge-preservation multi-classifier relearning framework for the
  classification of high-resolution remotely sensed imagery. ISPRS J.
  Photogramm. Remote Sens. 138, 57--73.

\bibitem[{Hang et~al.(2019)Hang, Liu, Hong, and Ghamisi}]{hang2019cascaded}
Hang, R., Liu, Q., Hong, D., Ghamisi, P., 2019. Cascaded recurrent neural
  networks for hyperspectral image classification. IEEE Trans. Geosci. Remote
  Sens. 57~(8), 5384--5394.

\bibitem[{Hardoon et~al.(2004)Hardoon, Szedmak, and
  J.~Shawe-Taylor}]{hardoon2004canonical}
Hardoon, D., Szedmak, S., J.~Shawe-Taylor, J., 2004. Canonical correlation
  analysis: An overview with application to learning methods. Neural Comput.
  16~(12), 2639--2664.

\bibitem[{Hong(2019)}]{hong2019regression}
Hong, D., 2019. Regression-induced representation learning and its optimizer: A
  novel paradigm to revisit hyperspectral imagery analysis. Ph.D. thesis,
  Technische Universit{\"a}t M{\"u}nchen.

\bibitem[{Hong et~al.(2020{\natexlab{a}})Hong, Chanussot, Yokoya, Kang, and
  Zhu}]{hong2020learning}
Hong, D., Chanussot, J., Yokoya, N., Kang, J., Zhu, X., 2020{\natexlab{a}}.
  Learning shared cross-modality representation using multispectral-lidar and
  hyperspectral data. IEEE Geosci. Remote Sens. Lett.DOI:
  10.1109/LGRS.2019.2944599.

\bibitem[{Hong et~al.(2015)Hong, Liu, Su, Pan, and Wang}]{hong2015novel}
Hong, D., Liu, W., Su, J., Pan, Z., Wang, G., 2015. A novel hierarchical
  approach for multispectral palmprint recognition. Neurocomputing 151,
  511--521.

\bibitem[{Hong et~al.(2020{\natexlab{b}})Hong, Wu, Ghamisi, Chanussot, Yokoya,
  and Zhu}]{hong2020invariant}
Hong, D., Wu, X., Ghamisi, P., Chanussot, J., Yokoya, N., Zhu, X.,
  2020{\natexlab{b}}. Invariant attribute profiles: A spatial-frequency joint
  feature extractor for hyperspectral image classification. IEEE Trans. Geosci.
  Remote Sens. 58~(6), 3791--3808.

\bibitem[{Hong et~al.(2019{\natexlab{a}})Hong, Yokoya, Chanussot, Xu, and
  Zhu}]{hong2019learning}
Hong, D., Yokoya, N., Chanussot, J., Xu, J., Zhu, X.~X., 2019{\natexlab{a}}.
  Learning to propagate labels on graphs: An iterative multitask regression
  framework for semi-supervised hyperspectral dimensionality reduction. ISPRS
  J. Photogramm. Remote Sens. 158, 35--49.

\bibitem[{Hong et~al.(2019{\natexlab{b}})Hong, Yokoya, Chanussot, and
  Zhu}]{hong2019augmented}
Hong, D., Yokoya, N., Chanussot, J., Zhu, X., 2019{\natexlab{b}}. An augmented
  linear mixing model to address spectral variability for hyperspectral
  unmixing. IEEE Trans. Image Process. 28~(4), 1923--1938.

\bibitem[{Hong et~al.(2019{\natexlab{c}})Hong, Yokoya, Chanussot, and
  Zhu}]{hong2019cospace}
Hong, D., Yokoya, N., Chanussot, J., Zhu, X.~X., 2019{\natexlab{c}}. Co{S}pace:
  Common subspace learning from hyperspectral-multispectral correspondences.
  IEEE Trans. Geosci. Remote Sens. 57~(7), 4349--4359.

\bibitem[{Hong et~al.(2019{\natexlab{d}})Hong, Yokoya, Ge, Chanussot, and
  Zhu}]{hong2019learnable}
Hong, D., Yokoya, N., Ge, N., Chanussot, J., Zhu, X., 2019{\natexlab{d}}.
  Learnable manifold alignment ({L}e{MA}): A semi-supervised cross-modality
  learning framework for land cover and land use classification. ISPRS J.
  Photogramm. Remote Sens. 147, 193--205.

\bibitem[{Hong et~al.(2017)Hong, Yokoya, and Zhu}]{hong2017learning}
Hong, D., Yokoya, N., Zhu, X., Jul 2017. Learning a robust local manifold
  representation for hyperspectral dimensionality reduction. IEEE J. Sel.
  Topics Appl. Earth Observ. Remote Sens. 10~(6), 2960--2975.

\bibitem[{Hong and Zhu(2018)}]{hong2018sulora}
Hong, D., Zhu, X., 2018. S{UL}o{RA}: Subspace unmixing with low-rank attribute
  embedding for hyperspectral data analysis. IEEE J. Sel. Topics Signal
  Process. 12~(6), 1351--1363.

\bibitem[{Hu et~al.(2019{\natexlab{a}})Hu, Hong, Wang, and
  Zhu}]{hu2019comparative}
Hu, J., Hong, D., Wang, Y., Zhu, X., 2019{\natexlab{a}}. A comparative review
  of manifold learning techniques for hyperspectral and polarimetric sar image
  fusion. Remote Sens. 11~(6), 681.

\bibitem[{Hu et~al.(2019{\natexlab{b}})Hu, Hong, and Zhu}]{Hu2019mima}
Hu, J., Hong, D., Zhu, X., 2019{\natexlab{b}}. M{IMA}: Mapper-induced manifold
  alignment for semi-supervised fusion of optical image and polarimetric sar
  data. IEEE Trans. Geosci. Remote Sens. 57~(11), 9025--9040.

\bibitem[{Ioffe and Szegedy(2015)}]{ioffe2015batch}
Ioffe, S., Szegedy, C., 2015. Batch normalization: Accelerating deep network
  training by reducing internal covariate shift. arXiv:1502.03167.

\bibitem[{Kampffmeyer et~al.(2016)Kampffmeyer, Salberg, and
  Jenssen}]{kampffmeyer2016semantic}
Kampffmeyer, M., Salberg, A., Jenssen, R., 2016. Semantic segmentation of small
  objects and modeling of uncertainty in urban remote sensing images using deep
  convolutional neural networks. In: Proc. CVPR Workshop. pp. 1--9.

\bibitem[{Kang et~al.(2020)Kang, Hong, Liu, Baier, Yokoya, and
  Demir}]{kang2020learning}
Kang, J., Hong, D., Liu, J., Baier, G., Yokoya, N., Demir, B., 2020. Learning
  convolutional sparse coding on complex domain for interferometric phase
  restoration. IEEE Trans. Neural Netw. Learn. Syst.DOI:
  10.1109/TNNLS.2020.2979546.

\bibitem[{Krizhevsky et~al.(2012)Krizhevsky, Sutskever, and
  Hinton}]{krizhevsky2012imagenet}
Krizhevsky, A., Sutskever, I., Hinton, G., 2012. Imagenet classification with
  deep convolutional neural networks. In: Proc. NIPS. pp. 1097--1105.

\bibitem[{Lanaras et~al.(2015)Lanaras, Baltsavias, and
  Schindler}]{lanaras2015hyperspectral}
Lanaras, C., Baltsavias, E., Schindler, K., 2015. Hyperspectral
  super-resolution by coupled spectral unmixing. In: Proc. ICCV. pp.
  3586--3594.

\bibitem[{LeCun et~al.(2015)LeCun, Bengio, and Hinton}]{lecun2015deep}
LeCun, Y., Bengio, Y., Hinton, G., 2015. Deep learning. Nature 521~(7553), 436.

\bibitem[{Li et~al.(2017)Li, Jie, Wang, Liu, Yang, Shen, Lin, Chen, Yan, and
  Feng}]{li2017foveanet}
Li, X., Jie, Z., Wang, W., Liu, C., Yang, J., Shen, X., Lin, Z., Chen, Q., Yan,
  S., Feng, J., 2017. Foveanet: Perspective-aware urban scene parsing. In:
  Proc. ICCV. pp. 784--792.

\bibitem[{Liu et~al.(2019)Liu, Deng, Chanussot, Hong, and Zhao}]{liu2019stfnet}
Liu, X., Deng, C., Chanussot, J., Hong, D., Zhao, B., 2019. Stfnet: A
  two-stream convolutional neural network for spatiotemporal image fusion. IEEE
  Trans. Geosci. Remote Sens. 57~(9), 6552--6564.

\bibitem[{Long et~al.(2015)Long, Shelhamer, and Darrell}]{long2015fully}
Long, J., Shelhamer, E., Darrell, T., 2015. Fully convolutional networks for
  semantic segmentation. In: Proc. CVPR. pp. 3431--3440.

\bibitem[{Luo et~al.(2017)Luo, Zou, Hoffman, and Fei-Fei}]{luo2017label}
Luo, Z., Zou, Y., Hoffman, J., Fei-Fei, L., 2017. Label efficient learning of
  transferable representations acrosss domains and tasks. In: Proc. NIPS. pp.
  165--177.

\bibitem[{Marcos et~al.(2018)Marcos, Tuia, Kellenberger, Zhang, Bai, Liao, and
  Urtasun}]{marcos2018learning}
Marcos, D., Tuia, D., Kellenberger, B., Zhang, L., Bai, M., Liao, R., Urtasun,
  R., 2018. Learning deep structured active contours end-to-end. In: Proc.
  CVPR. pp. 8877--8885.

\bibitem[{M{\'a}ttyus et~al.(2016)M{\'a}ttyus, Wang, Fidler, and
  Urtasun}]{mattyus2016hd}
M{\'a}ttyus, G., Wang, S., Fidler, S., Urtasun, R., 2016. Hd maps: Fine-grained
  road segmentation by parsing ground and aerial images. In: Proc. ICCV. pp.
  3611--3619.

\bibitem[{Melis et~al.(2017)Melis, Demontis, Biggio, Brown, Fumera, and
  Roli}]{melis2017deep}
Melis, M., Demontis, A., Biggio, B., Brown, G., Fumera, G., Roli, F., 2017. Is
  deep learning safe for robot vision? adversarial examples against the icub
  humanoid. In: Proc. ICCV. pp. 751--759.

\bibitem[{Ngiam et~al.(2011)Ngiam, Khosla, Kim, Nam, Lee, and
  Ng}]{ngiam2011multimodal}
Ngiam, J., Khosla, A., Kim, M., Nam, J., Lee, H., Ng, A., 2011. Multimodal deep
  learning. In: Proc. ICML. pp. 689--696.

\bibitem[{Nie et~al.(2018)Nie, Feng, and Yan}]{nie2018mutual}
Nie, X., Feng, J., Yan, S., 2018. Mutual learning to adapt for joint human
  parsing and pose estimation. In: Proc. ECCV. pp. 502--517.

\bibitem[{Noh et~al.(2015)Noh, Hong, and Han}]{noh2015learning}
Noh, H., Hong, S., Han, B., 2015. Learning deconvolution network for semantic
  segmentation. In: Proc. ICCV. pp. 1520--1528.

\bibitem[{Ouyang et~al.(2014)Ouyang, Chu, and Wang}]{ouyang2014multi}
Ouyang, W., Chu, X., Wang, X., 2014. Multi-source deep learning for human pose
  estimation. In: Proc. CVPR. pp. 2329--2336.

\bibitem[{Pal and Mitra(1992)}]{pal1992multilayer}
Pal, S., Mitra, S., 1992. Multilayer perceptron, fuzzy sets, and
  classification. IEEE Trans. Neural Netw. 3~(5), 683--697.

\bibitem[{Peng et~al.(2016)Peng, Huang, and Qi}]{peng2016cross}
Peng, Y., Huang, X., Qi, J., 2016. Cross-media shared representation by
  hierarchical learning with multiple deep networks. In: Proc. IJCAI. pp.
  3846--3853.

\bibitem[{Rastegar et~al.(2016)Rastegar, Soleymani, Rabiee, and
  Shojaee}]{rastegar2016mdl}
Rastegar, S., Soleymani, M., Rabiee, H., Shojaee, S.~M., 2016. M{DL}-{CW}: A
  multimodal deep learning framework with cross weights. In: Proc. CVPR. pp.
  2601--2609.

\bibitem[{Rasti et~al.(2020)Rasti, Hong, Hang, Ghamisi, Kang, Chanussot, and
  Benediktsson}]{rasti2020feature}
Rasti, B., Hong, D., Hang, R., Ghamisi, P., Kang, X., Chanussot, J.,
  Benediktsson, J., 2020. Feature extraction for hyperspectral imagery: The
  evolution from shallow to deep (overview and toolbox). IEEE Geosci. Remote
  Sens. Mag.DOI: 10.1109/MGRS.2020.2979764.

\bibitem[{Riese et~al.(2020)Riese, Keller, and Hinz}]{riese2020supervised}
Riese, F., Keller, S., Hinz, S., 2020. Supervised and semi-supervised
  self-organizing maps for regression and classification focusing on
  hyperspectral data. Remote Sens. 12~(1), 7.

\bibitem[{Silberer et~al.(2017)Silberer, Ferrari, and
  Lapata}]{silberer2017visually}
Silberer, C., Ferrari, V., Lapata, M., 2017. Visually grounded meaning
  representations. IEEE Trans. Pattern Anal. Mach. Intell. 39~(11), 2284--2297.

\bibitem[{Silberer and Lapata(2014)}]{silberer2014learning}
Silberer, C., Lapata, M., 2014. Learning grounded meaning representations with
  autoencoders. In: Proc. ACL. Vol.~1. pp. 721--732.

\bibitem[{Srivastava et~al.(2014)Srivastava, Hinton, Krizhevsky, Sutskever, and
  Salakhutdinov}]{srivastava2014dropout}
Srivastava, N., Hinton, G., Krizhevsky, A., Sutskever, I., Salakhutdinov, R.,
  2014. Dropout: a simple way to prevent neural networks from overfitting. J.
  Mach. Learn. Res. 15~(1), 1929--1958.

\bibitem[{Srivastava and
  Salakhutdinov(2012{\natexlab{a}})}]{srivastava2012learning}
Srivastava, N., Salakhutdinov, R., 2012{\natexlab{a}}. Learning representations
  for multimodal data with deep belief nets. In: Proc. ICML Workshop. Vol.~79.

\bibitem[{Srivastava and
  Salakhutdinov(2012{\natexlab{b}})}]{srivastava2012multimodal}
Srivastava, N., Salakhutdinov, R., 2012{\natexlab{b}}. Multimodal learning with
  deep boltzmann machines. In: Proc. NIPS. pp. 2222--2230.

\bibitem[{Srivastava et~al.(2019)Srivastava, Vargas-Mu{\~n}oz, and
  Tuia}]{srivastava2019understanding}
Srivastava, S., Vargas-Mu{\~n}oz, J., Tuia, D., 2019. Understanding urban
  landuse from the above and ground perspectives: A deep learning, multimodal
  solution. Remote Sens. Environ. 228, 129--143.

\bibitem[{Szegedy et~al.(2013)Szegedy, Zaremba, Sutskever, Bruna, Erhan,
  Goodfellow, and Fergus}]{szegedy2013intriguing}
Szegedy, C., Zaremba, W., Sutskever, I., Bruna, J., Erhan, D., Goodfellow, I.,
  Fergus, R., 2013. Intriguing properties of neural networks. arXiv:1312.6199.

\bibitem[{Tuia et~al.(2015)Tuia, Flamary, and Courty}]{tuia2015multiclass}
Tuia, D., Flamary, R., Courty, N., 2015. Multiclass feature learning for
  hyperspectral image classification: Sparse and hierarchical solutions. ISPRS
  J. Photogramm. Remote Sens. 105, 272--285.

\bibitem[{Tuia et~al.(2014)Tuia, Volpi, Trolliet, and
  Camps-Valls}]{tuia2014semisupervised}
Tuia, D., Volpi, M., Trolliet, M., Camps-Valls, G., 2014. Semisupervised
  manifold alignment of multimodal remote sensing images. IEEE Trans. Geosci.
  Remote Sens. 52~(12), 7708--7720.

\bibitem[{Vendrov et~al.(2015)Vendrov, Kiros, Fidler, and
  Urtasun}]{vendrov2015order}
Vendrov, I., Kiros, R., Fidler, S., Urtasun, R., 2015. Order-embeddings of
  images and language. arXiv:1511.06361.

\bibitem[{Wang et~al.(2014)Wang, Ooi, Yang, Zhang, and
  Zhuang}]{wang2014effective}
Wang, W., Ooi, B.~C., Yang, X., Zhang, D., Zhuang, Y., 2014. Effective
  multi-modal retrieval based on stacked auto-encoders. Proc. VLDB 7~(8),
  649--660.

\bibitem[{Wu et~al.(2020)Wu, Hong, Chanussot, Xu, Tao, and
  Wang}]{wu2019fourier}
Wu, X., Hong, D., Chanussot, J., Xu, Y., Tao, R., Wang, Y., 2020. Fourier-based
  rotation-invariant feature boosting: An efficient framework for geospatial
  object detection. IEEE Geosci. Remote Sens. Lett. 17~(2), 302--306.

\bibitem[{Wu et~al.(2019)Wu, Hong, Tian, Chanussot, Li, and Tao}]{wu2019orsim}
Wu, X., Hong, D., Tian, J., Chanussot, J., Li, W., Tao, R., 2019. O{RSI}m
  {D}etector: A novel object detection framework in optical remote sensing
  imagery using spatial-frequency channel features. IEEE Trans. Geosci. Remote
  Sens. 57~(7), 5146--5158.

\bibitem[{Xia et~al.(2016)Xia, Wang, Chen, and Yuille}]{xia2016zoom}
Xia, F., Wang, P., Chen, L., Yuille, A.~L., 2016. Zoom better to see clearer:
  Human and object parsing with hierarchical auto-zoom net. In: Proc. ECCV.
  Springer, pp. 648--663.

\bibitem[{Xia et~al.(2018)Xia, Bai, Ding, Zhu, Belongie, Luo, Datcu, Pelillo,
  and Zhang}]{xia2018dota}
Xia, G., Bai, X., Ding, J., Zhu, Z., Belongie, S., Luo, J., Datcu, M., Pelillo,
  M., Zhang, L., 2018. Dota: A large-scale dataset for object detection in
  aerial images. In: Proc. CVPR.

\bibitem[{Yamaguchi et~al.(2005)Yamaguchi, Moriyama, Ishido, and
  Yamada}]{yamaguchi2005four}
Yamaguchi, Y., Moriyama, T., Ishido, M., Yamada, H., 2005. Four-component
  scattering model for polarimetric sar image decomposition. IEEE Trans.
  Geosci. Remote Sens. 43~(8), 1699--1706.

\bibitem[{Yang et~al.(2019)Yang, Rosenhahn, and Murino}]{yang2019introduction}
Yang, M., Rosenhahn, B., Murino, V., 2019. Introduction to multimodal scene
  understanding. In: Multimodal Scene Understanding. Elsevier, pp. 1--7.

\bibitem[{Yao et~al.(2019)Yao, Meng, Zhao, Cao, and Xu}]{yao2019nonconvex}
Yao, J., Meng, D., Zhao, Q., Cao, W., Xu, Z., 2019. Nonconvex-sparsity and
  nonlocal-smoothness-based blind hyperspectral unmixing. IEEE Trans. Image
  Process. 28~(6), 2991--3006.

\bibitem[{Yu and Koltun(2015)}]{yu2015multi}
Yu, F., Koltun, V., 2015. Multi-scale context aggregation by dilated
  convolutions. arXiv:1511.07122.

\bibitem[{Yu et~al.(2019)Yu, Davis, and Fritz}]{yu2019attributing}
Yu, N., Davis, L., Fritz, M., 2019. Attributing fake images to gans: Learning
  and analyzing gan fingerprints. In: Proc. ICCV. pp. 7556--7566.

\bibitem[{Zampieri et~al.(2018)Zampieri, Charpiat, Girard, and
  Tarabalka}]{zampieri2018multimodal}
Zampieri, A., Charpiat, G., Girard, N., Tarabalka, Y., 2018. Multimodal image
  alignment through a multiscale chain of neural networks with application to
  remote sensing. In: Proc. ECCV.

\bibitem[{Zhang et~al.(2019{\natexlab{a}})Zhang, Zhang, Kang, Hong, Xu, and
  Zhu}]{zhang2019estimation}
Zhang, B., Zhang, M., Kang, J., Hong, D., Xu, J., Zhu, X., 2019{\natexlab{a}}.
  Estimation of pmx concentrations from landsat 8 oli images based on a
  multilayer perceptron neural network. Remote Sens. 11~(6), 646.

\bibitem[{Zhang et~al.(2018{\natexlab{a}})Zhang, Dana, Shi, Zhang, Wang, Tyagi,
  and Agrawal}]{zhang2018context}
Zhang, H., Dana, K., Shi, J., Zhang, Z., Wang, X., Tyagi, A., Agrawal, A.,
  2018{\natexlab{a}}. Context encoding for semantic segmentation. In: Proc.
  CVPR.

\bibitem[{Zhang et~al.(2019{\natexlab{b}})Zhang, Vosselman, Gerke, Persello,
  Tuia, and Yang}]{zhang2019detecting}
Zhang, Z., Vosselman, G., Gerke, M., Persello, C., Tuia, D., Yang, M.,
  2019{\natexlab{b}}. Detecting building changes between airborne laser
  scanning and photogrammetric data. Remote Sens. 11~(20), 2417.

\bibitem[{Zhang et~al.(2018{\natexlab{b}})Zhang, Vosselman, Gerke, Tuia, and
  Yang}]{zhang2018change}
Zhang, Z., Vosselman, G., Gerke, M., Tuia, D., Yang, M., 2018{\natexlab{b}}.
  Change detection between multimodal remote sensing data using siamese cnn.
  arXiv preprint arXiv:1807.09562.

\bibitem[{Zhao et~al.(2019)Zhao, Sveinsson, Ulfarsson, and
  Chanussot}]{zhao2019semi}
Zhao, B., Sveinsson, J., Ulfarsson, M., Chanussot, J., 2019. (semi-) supervised
  mixtures of factor analyzers and deep mixtures of factor analyzers
  dimensionality reduction algorithms for hyperspectral images classification.
  In: Proc. IGARSS. IEEE, pp. 887--890.

\bibitem[{Zhao et~al.(2017)Zhao, Shi, Qi, Wang, and Jia}]{zhao2017pyramid}
Zhao, H., Shi, J., Qi, X., Wang, X., Jia, J., 2017. Pyramid scene parsing
  network. In: Proc. CVPR. pp. 2881--2890.

\bibitem[{Zhu et~al.(2005)Zhu, Lafferty, and Rosenfeld}]{zhu2005semi}
Zhu, X., Lafferty, J., Rosenfeld, R., 2005. Semi-supervised learning with
  graphs. Ph.D. thesis, Carnegie Mellon University, Language Technologies
  Institute, School of Computer Science.

\end{thebibliography}
\end{document}